\documentclass[runningheads]{llncs}
\usepackage{graphicx}
\usepackage{tikz}
\usepackage{comment}
\usepackage{amsmath,amssymb} % define this before the line numbering.
\usepackage{color}

\usepackage[accsupp]{axessibility}  % Improves PDF readability for those with disabilities.

\usepackage{orcidlink}
\usepackage{booktabs,paralist}
\usepackage{threeparttable}
\usepackage{adjustbox}

\usepackage{epsfig}
\usepackage{array}
\usepackage{multicol, multirow}
\usepackage{vwcol}
\graphicspath{{figures/}{Figures/}}
\usepackage{wrapfig,bm}

\DeclareMathOperator*{\argmin}{arg\,min}
\usepackage{colortbl, array}
\usepackage{hhline}
\usepackage{booktabs}

\usepackage{amsmath,amssymb,comment,tabularx}       % hyperlinks
\usepackage{url}            % simple URL typesetting
\usepackage{booktabs}       % professional-quality tables
\usepackage{nicefrac}       % compact symbols for 1/2, etc.
\usepackage{microtype}      % microtypography
\usepackage[linesnumbered,ruled]{algorithm2e}
\usepackage{caption,color}
\usepackage{accents}
\usepackage{comment}
\usepackage[switch]{lineno}

\usepackage{algorithmic}
\usepackage{mathrsfs}

\usepackage[capitalize]{cleveref}
\crefname{section}{Sec.}{Secs.}
\Crefname{section}{Section}{Sections}
\Crefname{table}{Table}{Tables}
\crefname{table}{Tab.}{Tabs.}

\definecolor{rulecolor}{RGB}{0,71,171}
\definecolor{tableheadcolor}{RGB}{204,229,255}
\definecolor{Gray}{gray}{0.9}

\newcommand{\myrowcolour}{\rowcolor{tableheadcolor}}

\newcommand{\topline}{ %
        \arrayrulecolor{rulecolor}\specialrule{0.1em}{\abovetopsep}{0pt}%
        \arrayrulecolor{rulecolor}}
\newcommand{\midtopline}{ %
        \arrayrulecolor{tableheadcolor}\specialrule{\aboverulesep}{0pt}{0pt}%
        \arrayrulecolor{rulecolor}\specialrule{\lightrulewidth}{0pt}{0pt}%
        \arrayrulecolor{white}\specialrule{\belowrulesep}{0pt}{0pt}%
        \arrayrulecolor{rulecolor}}
\newcommand{\bottomline}{ %
        \arrayrulecolor{rulecolor} %
        \specialrule{\heavyrulewidth}{0pt}{\belowbottomsep}}%

\newtheorem{assumption}{Assumption}
\newcolumntype{g}{>{\columncolor{Gray}}c}

\begin{document}
% \linenumbers
\pagestyle{headings}
\mainmatter
\def\ECCVSubNumber{3213}  % Insert your submission number here

\title{On the Versatile Uses of Partial Distance Correlation in Deep Learning} % Replace with your title

\begin{comment}
\titlerunning{ECCV-22 submission ID \ECCVSubNumber} 
\authorrunning{ECCV-22 submission ID \ECCVSubNumber} 
\author{Anonymous ECCV submission}
\institute{Paper ID \ECCVSubNumber}
\end{comment}
%******************

% \begin{comment}
\titlerunning{On the Versatile Uses of Partial Distance Correlation in Deep Learning}
\author{Xingjian Zhen\inst{1}\orcidlink{0000-0002-2220-8246}  \and
Zihang Meng\inst{1}\orcidlink{0000-0002-1919-0591} \and
Rudrasis Chakraborty\inst{2}\orcidlink{0000-0002-0448-911X} \and Vikas Singh\inst{1}\orcidlink{0000-0002-8355-6519}}
\authorrunning{X. Zhen et al.}
\institute{University of Wisconsin-Madison \\ {\bf \{xzhen3, zmeng29\}@wisc.edu, vsingh@biostat.wisc.edu} \and Butlr \\
{\bf rudrasischa@gmail.com}
}
% \end{comment}
%******************
\maketitle

%%%%%%%%% ABSTRACT
\begin{abstract}
   Comparing the functional behavior of neural network models, whether it is a single network over time or two (or more networks) during or post-training, is an essential step in understanding what they are learning (and what they are not), and for identifying strategies for regularization or efficiency improvements. Despite recent progress, e.g., comparing vision transformers to CNNs, systematic comparison of function, especially across different networks, remains difficult and is often carried out layer by layer. Approaches such as canonical correlation analysis (CCA) are applicable in principle, but have been sparingly used so far. In this paper, we revisit a (less widely known) from statistics, called distance correlation (and its partial variant), designed to evaluate correlation between feature spaces of different dimensions. We describe the steps necessary to carry out its deployment for large scale models -- this opens the door to a surprising array of applications ranging from conditioning one deep model w.r.t. another, learning disentangled representations as well as optimizing diverse models that would directly be more robust to adversarial attacks. Our experiments suggest a versatile regularizer (or constraint) with many advantages, which avoids some of the common difficulties one faces in such analyses
   \footnote[1]{Code is at   \url{https://github.com/zhenxingjian/Partial_Distance_Correlation}}.
\end{abstract}

%%%%%%%%% BODY TEXT
\section{Introduction}
\label{sec:intro}
% {\color{red}
% Key point, how to compare between two networks. Current tools are not super satisfactory. On one hand, 
% Do Vision Transformers See Like Convolutional Neural Networks? discussed about some aspect of this question. On another, we can compare the behavior of individual samples and Grad-CAM. 
% Another small aspect of this question is the invariant and fairness. How should we learn something unique.
% Controlling on one network and training another one fundamentally requires information theory. How hard this can be. We can go statistic route/ KL divergence, but it relies on the same stand. Other method like Deep CCA, has some more work to answer such question. Then, we're studying the DC and PDC. 
% }
%The early development of neural networks, 
%especially in vision, was heavily inspired 
%by biological vision and the contemporary neuroscientific 
%understanding at that time of 
%how the human eye acquired and processed 
%visual data. 
%https://arxiv.org/pdf/2001.07092.pdf
%The discovery of two types of simple cells 
%in the visual cortex of cats by Hubel and Wiesel 
%\cite{Hubel, D. H., & Wiesel, T. N. (1962). Receptive fields, binocular interaction and functional
%architecture in the cat's visual cortex. The Journal of physiology, 160(1), 106-154.} 
%eventually led to 
%Neocognitron,
%likely the first functional ``model'' of the
%visual system, 
%which some consider to be the ancestor 
%of convolutional neural networks (CNNs). 
The extent to which popular architectures in computer vision 
even partly mimic human vision continues to be studied (and debated) 
in our community. 
But consider the following hypothetical 
scenario. Let us say that a fully functional 
{\em computational} model of the visual system -- perhaps 
a modern version of the Neocognitron \cite{fukushima1983neocognitron} -- 
was somehow provided to us. And 
we wished to ``compare'' its 
behavior to modern CNN models \cite{iandola2014densenet,he2016deep}.  
To do so, two options appear sensible. 
The first -- inspired by analogies between 
computational vision and biological vision -- 
would draw a correspondence between how simple/complex cells 
in the visual cortex process scenes 
and their induced receptive fields with those of 
activations of units/blocks 
in a modern deep neural network architecture \cite{selvaraju2017grad}. 
While this process is often difficult to carry out systematically, 
it is powerful and, in some ways, has contributed to 
interest in biologically inspired deep learning, see \cite{wozniak2020deep}. 
Updated forms of this  
intuition -- associating different subsets of cells (or neural network units) to different semantic/visual concepts -- remains the default approach we use in debugging and interpretation. 
The second option for tackling the hypothetical setting above is 
to pose it in an information theoretic setting. That is, for 
two models $\Theta_X$ and $\Theta_Y$, we ask the following question: {\em what 
has $\Theta_X$ learned that $\Theta_Y$ has not? Or vice versa}. The asymmetry
is intentional because if we consider two random variables (r.v.) $X,Y$, 
the question simply takes the form of ``conditioning'', i.e., 
compare $\mathbb{P}(X)$ versus $\mathbb{P}(X|Y)$. This form suffices 
if our interest is restricted to the {\em predictions} of the two models. 
If we instead wish to capture 
the model's behavior more globally -- when $X$ and $Y$ denote 
the full set of feature responses -- we can use 
divergence measures on high dimensional probability 
measures given by 
the two models ($\Theta_X$ and $\Theta_Y$) responses on the training samples.  
Importantly, notice that our description assumes that, at least, the probability measures 
are defined on the same domain. 

\noindent {\bf More general use cases.} While the above discussion was cast as
comparing two networks, it is representative of a broad 
basket of tasks in deep learning. {\bf (a)} Consider the problem 
of learning fair representations \cite{zemel2013learning,feldman2015certifying,zafar2017fairness,lokhande2020fairalm} where the model must be invariant 
to one (or more) sensitive attributes. We seek latent representations, say $\Psi_{\rm pred}(X)$ for 
the prediction task, which 
minimizes mutual information w.r.t. the latent representation 
relevant for predicting the sensitive attribute $\Psi_{\rm sens}(X)$. 
Indeed, if information regarding the sensitive attribute is partially preserved or leaks into $\Psi_{\rm pred}(X)$, the relative entropy will be low \cite{moyer2018invariant}. 
Observe that this calculation is possible partly 
because the latent space specifies 
the {\em same probability space} for the two distributions. {\bf (b)} The setting is 
identical in common approaches for learning disentangled representations, 
where disentanglement is measured via various information theoretic measures \cite{chen2018isolating,achille2018emergence,gabbay2021image,shu2019weakly}.  
If we now segue back to comparing two different networks, but without the convenience of 
a common coordinate system to measure divergence, the options turn out to be 
 limited. {\bf (c)} Recently, in trying to understand whether vision Transformers ``see'' similar to convolutional neural
networks \cite{raghu2021vision}, one option utilized 
recently was a kernel-based representation similarity, in a layer-by-layer manner. What 
we may actually want is a mechanism for conditioning -- for example, if one of the models is 
thought of a ``nuisance variable'', we wish to check the residual 
in the other after the first has been controlled for (or marginalized out). Importantly, 
this should be possible without assuming that the probability distributions live 
in the same space (or networks $\Theta_X$ and $\Theta_Y$ are the same). 

\noindent{\bf A direct application of CCA?} Consider two different feature spaces $(\mathcal{X}$ and $\mathcal{Y})$, 
say in dimensions $\mathbb{R}^p$ and $\mathbb{R}^{q}$, pertaining to 
feature activations from two different models. 
Comparison of these two feature spaces {\em is}
possible. One natural choice is canonical correlation analysis (CCA) \cite{bach2005probabilistic}, a generalization 
of correlation, specifically suited 
when $p \neq {q}$. 
The idea has been utilized for studying representation similarity in deep neural network models \cite{morcos2018insights},  
albeit in a post-training setting for reasons that will be clear shortly, as well as for identifying 
more efficient training regimes (i.e., can lower layers be sequentially frozen after a
certain number of timesteps). CCA has also been shown to be implementable 
within DNN pipelines for multi-view training, called DeepCCA \cite{andrew2013deep}, although efficiency can be a 
bottleneck limiting its broader deployment. A stochastic version 
of CCA suitable for DNN training with mini-batches has been proposed 
very recently, 
and strong experimental evidence was presented \cite{meng2021online}, also see \cite{gemp2022generalized}. 
Given that a stochastic CCA is now available, its extensions to the partial CCA setting are not yet available. If successful, this may 
eventually provide a scheme, suitable for deep learning, for 
controlling 
the influence of one model (or a set of variables) on another model. 

\noindent{\bf This work.} The starting 
point of this work is a less widely used 
statistical concept to measure 
the correlation between two different feature spaces $(\mathcal{X}, \mathcal{Y})$ of {\em different dimensions}, called distance correlation 
(and the method of dissimilarities). In shallow settings, CCA 
and distance correlation offers 
very similar functionality -- for the most part, they can be used interchangeably although distance correlation would {\em also 
need} specification of distances (or dissimilarities). In other words, 
CCA may be easier to deploy. On the other hand, deep variants of 
CCA involve specialized algorithms \cite{andrew2013deep,meng2021online}. Further, deep  
versions of partial CCA have not been reported. In contrast, 
as long as feature distances {\em can} be calculated, the differences 
between the shallow and the deep versions of distance correlation are minimal at best, and adjustments needed are quite minor. These advantages carry over 
to partial distance correlation, directly enabling conditioning 
one model w.r.t. another (or using such a term as a regularizer). 
The main {\bf contribution} of this paper is to 
study  
distance correlation (and partial distance correlation) as a powerful  
measure in a broad suite of tasks in vision. We review 
the relevant technical steps which enable its instantiation in 
deep learning settings and show its broad applications ranging 
from learning disentangled representations to understanding the 
differences between what two (or more) networks are learning to training 
``mutually distinct'' deep models (akin to earlier works 
on $M$ best solutions to MAP estimation in graphical models \cite{fromer2009lp,yadollahpour2011diverse}) 
or training $M$ diverse models for foreground-background segmentation as well as other tasks \cite{guzman2014efficiently}.

\subsection{Related Works}

Four distinct lines of work are related 
to our development, which 
we review next. 

\noindent{\bf Similarity between networks.} 
Understanding the similarity between 
different networks is an active topic \cite{kornblith2019better,geirhos2018imagenet,neyshabur2020being} also 
relevant in adversarial models  \cite{demontis2019adversarial,cheng2019improving}. 
Early attempts to compare neural network representations were approached via linear regression \cite{ramsay1984matrix}, whose 
applicability 
to nonlinear models is limited. 
As noted above, canonical correlation analysis (CCA) \cite{anderson1958introduction,hotelling1992relations} is a suitable 
off-the-shelf method for model comparisons. To this end, singular vector CCA (SVCCA) \cite{raghu2017svcca}, 
% which works on the truncated singular value decompositions of the random variables
Projection-Weighted CCA \cite{morcos2018insights}, 
DeepCCA \cite{andrew2013deep},
and stochastic CCA \cite{gao2019stochastic} are all potentially useful. 
Recently, \cite{kornblith2019similarity} 
studied the invariance properties for a good similarity measurement and proposed the centered kernel alignment (CKA). CKA offers invariance to invertible linear transformations, orthogonal transformations, and isotropic scaling. Separately, \cite{nguyen2020wide,raghu2021vision} used CKA to study similarities between deep and wide neural networks and also between different network structures. 

\noindent{\bf Information theoretic divergence measures.}
% Look at Aditya/Vishnu ICL paper. 
Another body of related work pertains to approximately measuring the mutual information \cite{cover1999elements} to remove this information, mainly in the 
context of fair representation learning.
Here, mutual information (MI) is measured 
between features and the sensitive attribute \cite{moyer2018invariant}. 
{
In \cite{song2019learning}, another information theoretic bound for learning maximally expressive representations subject to the given attributes is presented.
% using KL divergence between the prior distribution and the distribution conditioned on the given attributes. 
In \cite{cho2020fair}, MI between prediction and the sensitive attributes is used to train a fair classifier whereas \cite{akash2021learning} describes the use of inverse contrastive loss.
}
Group-theoretic 
approaches have also been described in \cite{cohen2016group,lokhande2022equivariance}. The work in \cite{lample2017fader} gives an empirical solution to remove specific visual features from the latent variables using adversarial training. 
% {\color{red} 1-2 more sentences?}

\noindent{\bf Repulsion/Diversity.} 
% Repulsive Deep Ensembles, Differentiable Quality Diversity.
If we consider the ensemble of neural networks, there are several different strategies to maintain 
functional diversity between ensemble members -- we 
acknowledge these results here because they are loosely related to one of the use cases we evaluate later. 
SVGD \cite{d2021stein} shows the benefits of choosing the kernel to measure the similarity between ensemble members. 
In \cite{d2021repulsive}, the authors introduce a kernelized repulsive term in the training loss, 
which endows deep ensembles with Bayesian convergence properties.
The so-called quality diversity (QD) is interesting:  \cite{pugh2016quality} tries to maximize a given objective function with diversity to a set of pre-defined measure functions \cite{gaier2020discovering,rakicevic2021policy}. When both the objective and measure functions in QD are differentiable, \cite{fontaine2021differentiable} offers an efficient way to explore the latent space of the objective w.r.t. the measure functions.

\noindent{\bf Distance correlation (DC).} The central idea 
motivating our work is distance correlation 
described in \cite{szekely2007measuring}. It has been used in the analysis of nonlinear dependence in time-series \cite{zhou2012measuring}, and feature screening in ultra high-dimensional data analysis tasks \cite{li2012feature}
and we will review it in detail shortly.

\section{Review: Distance (and Partial Distance) Correlation}
\label{sec:dc}
% {\color{red}
% Introducing DC \cite{szekely2007measuring} and PDC \cite{szekely2014partial}. Some toy examples. Why we need a statistical version. 1 page or more. Key is to show that DC is very important and powerful, and extending it to deep neural network will be beneficial.
% }

Given two random variables $X,Y \in \mathbb{R}$ (in the same domain), correlation (say, the Pearson correlation) helps measure their association. One can derive meaningful conclusions 
by statistical testing. 
% (true correlation 
% coefficient is zero) 
%and obtaining 
%confidence intervals.
% (specifying a probability that 
% the correlation coefficient lies in the interval). 
% A number of adjusted, weighted and scaled variants 
% are available and widely used. 
As noted in \S\ref{sec:intro}, one generalization of 
correlation to a higher dimension is CCA, which seeks to 
find projection matrices such 
that correlation among the projected data is maximized, see \cite{bach2005probabilistic}.  

%More recently, in
\noindent {\bf Benefits of Distance Correlation.} In many applications, the notion of distances or dissimilarities 
appears quite naturally. Motivated by the need for a scheme that 
can capture both linear and non-linear correlations when provided 
with such dissimilarity information, in 
\cite{szekely2007measuring}, the authors proposed a new measure of dependence between random vectors, called {\bf distance correlation}. The key benefits of distance correlation are: 
\begin{compactenum}
    \item The distance correlation $\mathcal{R}$ satisfies $0\leq \mathcal{R} \leq 1$, and $\mathcal{R}=0$ if and only if $X,Y$ are independent.
     \item  $\mathcal{R}(X,Y)$ is defined for $X$ and $Y$ in {\bf arbitrary dimensions}, e.g., $\mathcal{R}(X,Y)$ is well-defined when $X$ is of dimension $p$ while $Y$ is of dimension $q$ for $p \neq q$.
\end{compactenum}

% Here, we focus on empirical distance correlation 
% {\color{red}since our goal 
% is to compare two neural networks (i.e., representations of a finite dataset induced by the pair of networks).} -- no need to mention neural network here  
We focus on empirical distance correlation for $n$ samples drawn from the unknown joint distribution, 
and review its calculation.

For an observed random sample $(x,y) = \{(X_i,Y_i):i=1,\cdots, n\}$ from the joint distribution of random vectors $X$ in $\mathbb{R}^p$ and $Y$ in $\mathbb{R}^q$, define:
\vspace{-1em}
\begin{align}
    &a_{k,l} = \|X_k-X_l\|, \quad \bar{a}_{k, \cdot}=\frac{1}{n}\sum_{l=1}^n a_{k,l}, \quad \bar{a}_{\cdot, l}=\frac{1}{n} \sum_{k=1}^n a_{k,l}, \nonumber \\[-1em]
    &\bar{a}_{\cdot, \cdot} = \frac{1}{n^2} \sum_{k,l=1}^n a_{k,l}, \quad A_{k,l} = a_{k,l} - \bar{a}_{k,\cdot} -\bar{a}_{\cdot, l} + \bar{a}_{\cdot, \cdot} \label{eq:DC_A}
\end{align}
\vspace{-1em}

\noindent where $k,l \in \{1,\cdots,n\}$. Similarly, we can define $b_{k,l} = \|Y_k-Y_l\|$, and $B_{k,l}= b_{k,l} - \bar{b}_{k,\cdot} -\bar{b}_{\cdot, l} + \bar{b}_{\cdot, \cdot}$, and  
based on these quantities we have.

\vspace{-0.5em}
\begin{definition}
    {\it (Distance correlation) \cite{szekely2007measuring}.} The empirical distance correlation $\mathcal{R}_n(x, y)$ is the square root of 
    \vspace{-1em}
    \begin{align}
        \mathcal{R}_n^2(x, y) = \left\{
    \begin{array}{cl}
         \frac{\mathcal{V}_n^2(x, y)}{\sqrt{\mathcal{V}_n^2(x, x)\mathcal{V}_n^2(y, y)}}  &, \mathcal{V}_n^2(x, x)\mathcal{V}_n^2(y, y) > 0\\
        0 &, \mathcal{V}_n^2(x, x)\mathcal{V}_n^2(y, y) = 0
    \end{array}
    \right.
    \end{align}
    \vspace{-1em}
    
    \noindent where the empirical distance covariance (variance) $\mathcal{V}_n(x, y), \mathcal{V}_n(x, x)$ are defined as 
    $\mathcal{V}_n^2(x, y) = \frac{1}{n^2} \sum_{k,l=1}^n A_{k,l} B_{k,l},\mathcal{V}_n^2(x, x) = \frac{1}{n^2} \sum_{k,l=1}^n A_{k,l}^2$, with $A$ in \eqref{eq:DC_A}.
    \label{def:1}
\end{definition}
\vspace{-0.5em}

\begin{figure}[!b]
\vspace{-1em}
        \centering
               \includegraphics[width=0.23\columnwidth]{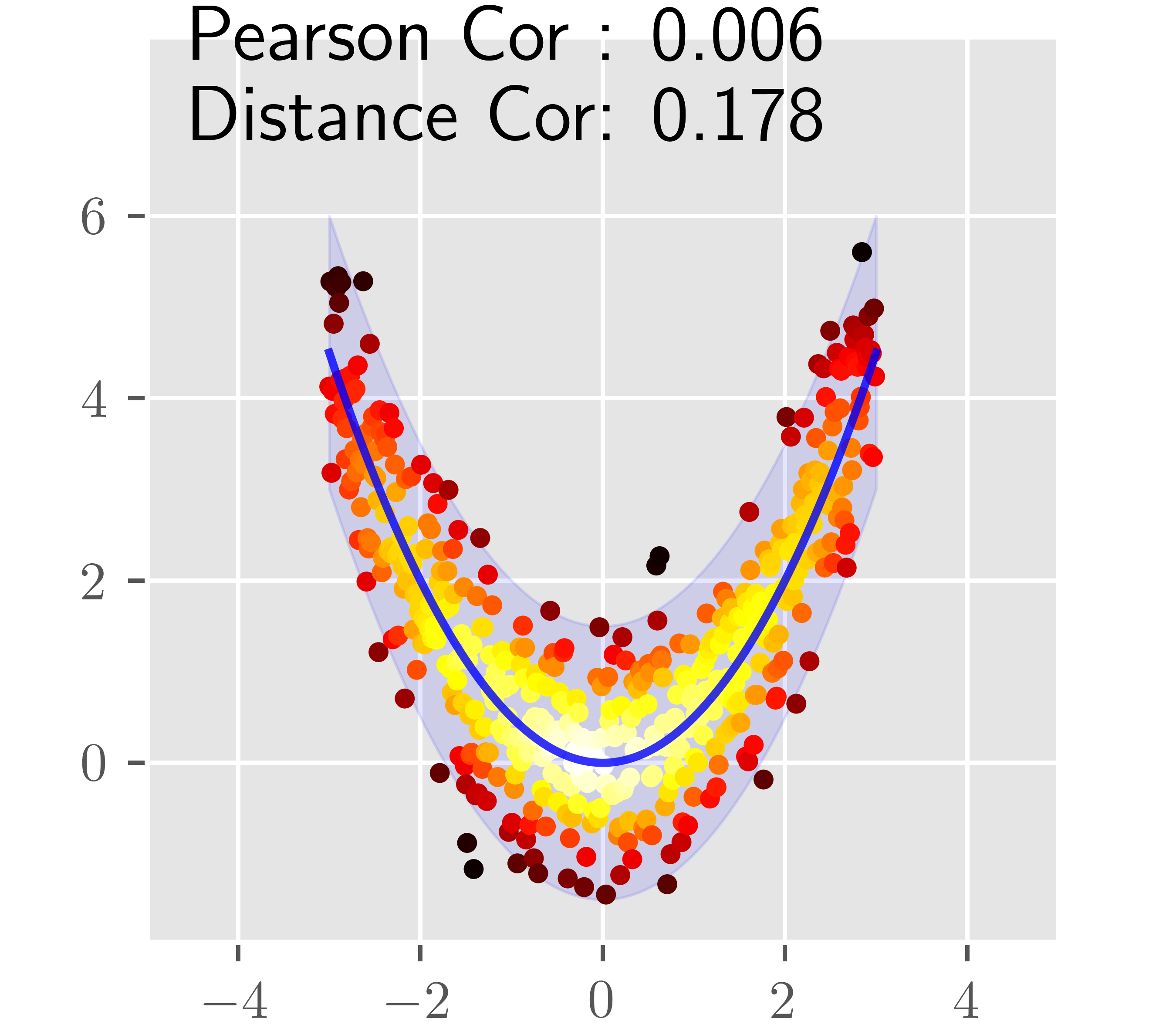}
                \includegraphics[width=0.23\columnwidth]{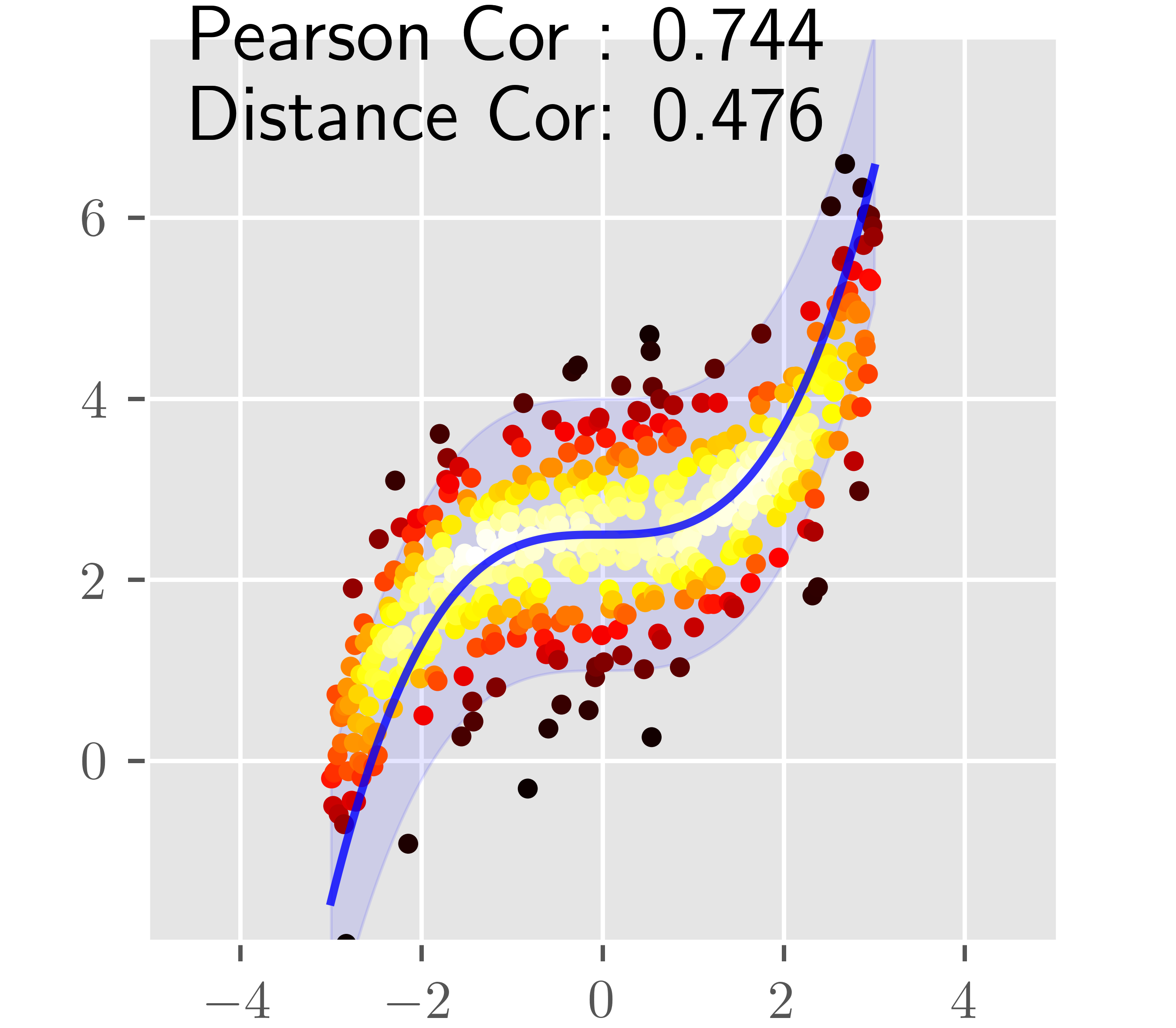}
               \includegraphics[width=0.23\columnwidth]{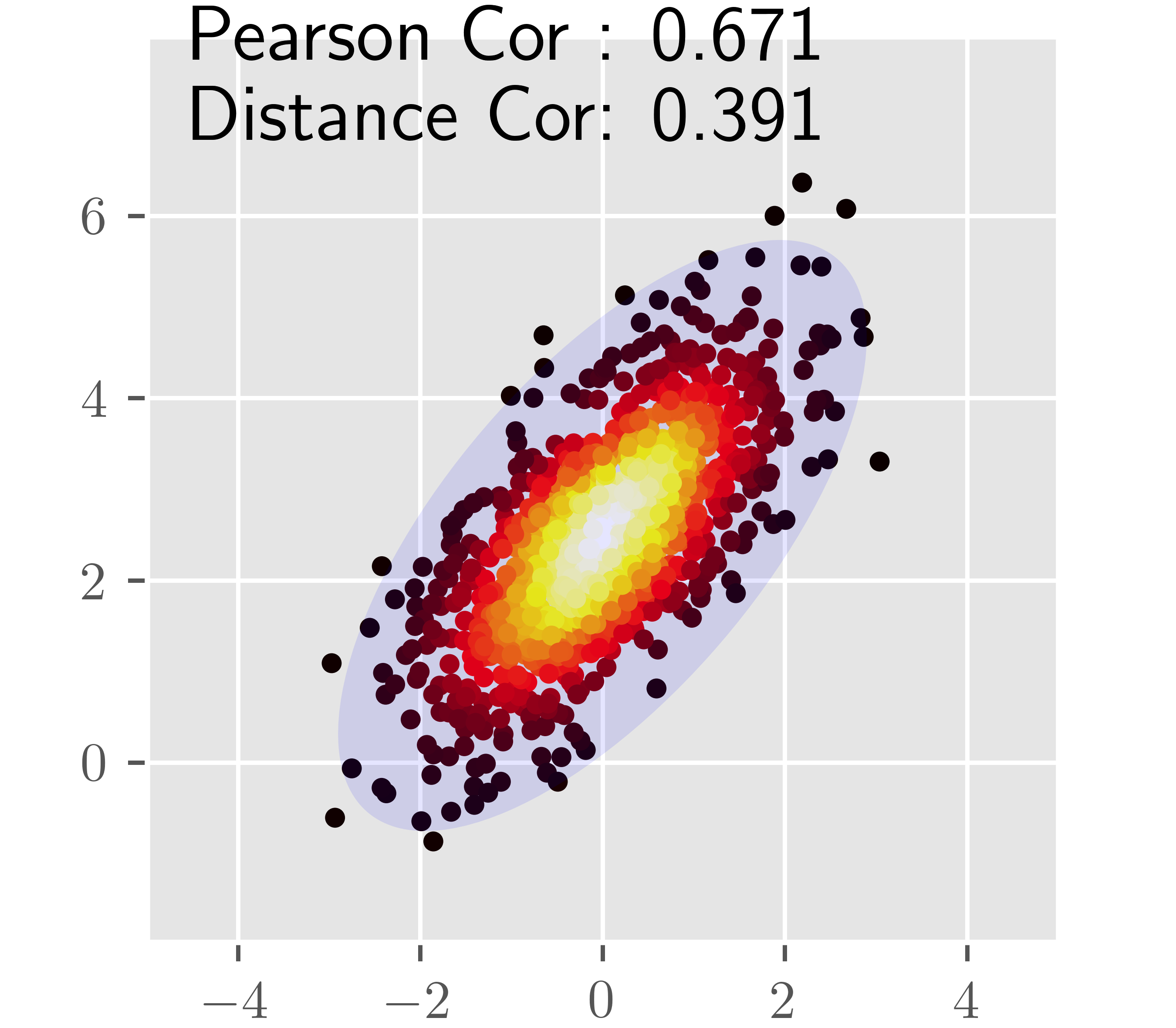}
               \includegraphics[width=0.23\columnwidth]{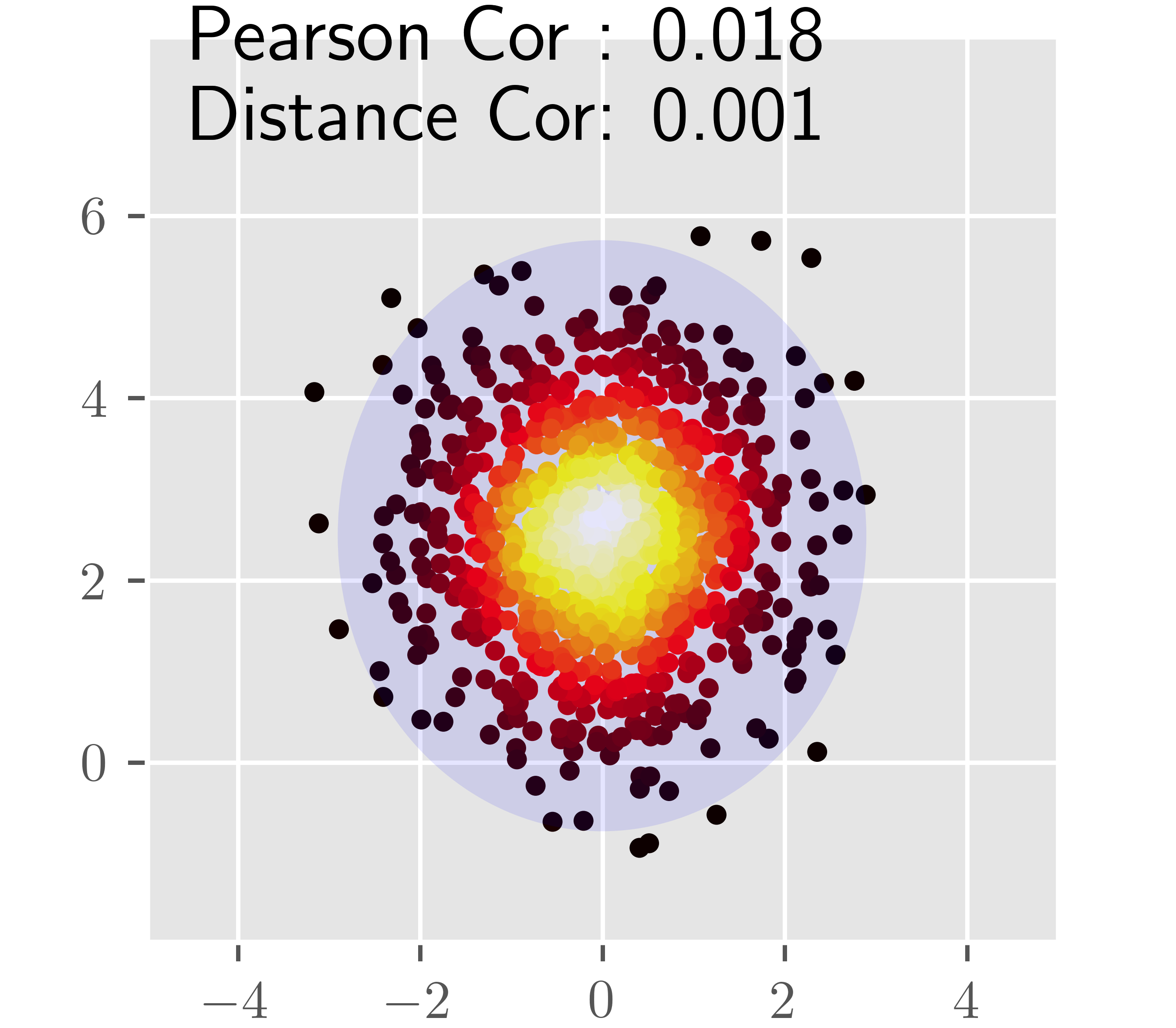}
\vspace{-1em}
               \caption{\footnotesize  Examples of Pearson Correlation and Distance Correlation in different settings. (a): $y=0.5x^2+0.75n,n\sim\mathcal{N}(0,1)$; (b): $y=0.15x^3+0.75n+2.5,n\sim\mathcal{N}(0,1)$; (c): $\begin{bmatrix} x \\ y \end{bmatrix} \sim \mathcal{N} \left(\begin{bmatrix} 0 \\ 2.5 \end{bmatrix},\begin{bmatrix} 1 & 0.75 \\ 0.75 & 1.25 \end{bmatrix}\right)$; (d): $\begin{bmatrix} x \\ y \end{bmatrix} \sim \mathcal{N}\left(\begin{bmatrix} 0 \\ 2.5 \end{bmatrix},\begin{bmatrix} 1 & 0 \\ 0 & 1.25 \end{bmatrix}\right)$}
               \label{fig:ex_pear_dist}
\end{figure}

\noindent {\bf Examples.} We show a few simple 2D examples to 
contrast Pearson Correlation and Distance Correlation in Fig. \ref{fig:ex_pear_dist}. 
Notice that if the relationship between 
the two random variables is not linear, Pearson Correlation might be small while Distance Correlation remains meaningful. 
% On the other hand, if $X$ and $Y$ are linearly independent, both correlations will be small. 

\noindent {\bf Extensions to 
conditioning.} Given three random variables $X$, $Y$, and $Z$, we want to measure the correlation between $X$ and $Y$ but ``controlling for'' $Z$ (thinking 
of it as a nuisance variable), i.e., 
we want to estimate $\mathcal{R}(X|Z, Y|Z)=\mathcal{R}^* (X,Y;Z)$.
Such a quantity is key in existing 
approaches in disentangled learning, deriving invariant 
representations and understanding what one or more networks are learning 
after concepts learned by another network have been accounted for. 
Consider how this task would be accomplished in linear regression. We 
would project $X$ and $Y$ into the space of $Z$, 
and only use the residuals to measure the correlation. 
%However, this is very difficult when the dimensions of $X$, $Y$, and $Z$ being different. 
% Distance correlation can easily 
% handle the dimensionality mismatch because {\em the quantity 
% is calculated using distances between pairs of samples in each space}. 
%difference. But the problem of 
Nonetheless, defining partial distance correlation is more involved -- in \cite{szekely2014partial}, the authors introduced a new Hilbert space where we can define the projection of distance matrix. % use an inner product on the squared distance covariance. 
%To begin with, we will need to modify a little to 
To do so, the authors calculate a 
$\mathcal{U}$-centered matrix $\tilde A$ from the distance 
matrix $(a_{k,l})$ so that the inner product of the $\mathcal{U}$-centered matrices will be the distance covariance. 

\begin{definition}
    Let $A = (a_{k,l})$ be a symmetric, real valued $n\times n$ matrix ($n>2$) with zero diagonal. Define the $\mathcal{U}$-centered matrix $\tilde A = (\tilde{a}_{kl})$ as follows.
    \vspace{-1em}
    {\small
    \begin{align}
        \tilde{a}_{kl} = \left\{
    \begin{array}{cl}\displaystyle
         a_{k,l} - \frac{1}{n-2}\sum_{i=1}^n a_{i,l} - \frac{1}{n-2}\sum_{j=1}^n a_{k,j} 
         + \frac{1}{(n-1)(n-2)}\sum_{i,j=1}^n a_{i,j} &, k\neq l\\
        0 &, k=l
    \end{array}
    \right.
    \end{align}
    }%
    
    %\vspace{-1em}
    \noindent Further, the inner product between $\tilde A, \tilde B$ is defined as 
    $
        (\tilde A \cdot \tilde B) := \frac{1}{n(n-3)} \sum_{k\neq l} \tilde A_{k,l} \tilde B_{k,l}
    $,
    and is an unbiased estimator of squared population distance covariance $\mathcal{V}^2(x,y)$. 
\end{definition}
\vspace{-0.5em}

% The above inner product, as noted in \cite{szekely2014partial}, is defined on the Hilbert space spanned by all $n\times n$ matrices $\tilde A$. 
Before defining partial distance covariance formally, we recall the definition of orthogonal projection on these matrices.

\vspace{-0.5em}
\begin{definition}
    Let $\tilde A, \tilde B, \tilde C$ corresponding to samples $x, y, z$ respectively, and let
    $
        P_{z^{\perp}}(x) = \tilde A - \frac{(\tilde A \cdot \tilde C)}{(\tilde C \cdot \tilde C)}\tilde C, \quad
        P_{z^{\perp}}(y) = \tilde B - \frac{(\tilde B \cdot \tilde C)}{(\tilde C \cdot \tilde C)}\tilde C
    $
    denote the orthogonal projection of $\tilde A(x)$ onto $(\tilde C(z))^{\perp}$ and the orthogonal projection of $\tilde B(y)$ onto $(\tilde C(z))^{\perp}$. 
\end{definition}
\vspace{-0.5em}

% Naturally, $P_{z^{\perp}}(x)$ and $P_{z^{\perp}}(y)$ also live in the same Hilbert space. 
Now, we are ready to define the partial distance covariance and the partial distance correlation. 

\vspace{-0.5em}
\begin{definition}
    Let (x,y,z) be a random sample observed from the joint distribution of $(X,Y,Z)$. The sample partial distance covariance is defined by:
    \vspace{-1em}
    \begin{align}
    \label{eq:pdcov}
        \text{pdCov}(x,y;z)&= (P_{z^{\perp}}(x) \cdot P_{z^{\perp}}(y))
                            = \frac{1}{n(n-3)}\sum_{i\neq j} \left(P_{z^{\perp}}(x) \right)_{i,j}\left(P_{z^{\perp}}(y)\right)_{i,j}
    \end{align}
    
    \vspace{-1em}
    \noindent And the partial distance correlation is defined as:
    $
        {\mathcal{R}^*}^2(x,y;z) := \frac{(P_{z^{\perp}}(x) \cdot P_{z^{\perp}}(y))}{\|P_{z^{\perp}}(x)\| \|P_{z^{\perp}}(y)\|}
    $
    where $\|P_{z^{\perp}}(x)\| = (P_{z^{\perp}}(x)\cdot P_{z^{\perp}}(x))^{1/2}$ is the norm.
\end{definition}
\vspace{-0.5em}

%The reader will notice that the expression for partial distance correlation can enable answering the question: {\it what does $X$ related to what $Y$ learns that $Z$ does not?} where $X$, $Z$ are the features learned from two neural networks, and $Y$ can be considered to be the known variables. 
Partial distance correlation enables asking various interesting questions. 
By projecting the original $\mathcal{U}$-centered matrix $\tilde A$ onto $\tilde C$, the correlation between the residual and $\tilde B$ will be a measure of what does $X$ learn that $Z$ does not. 

% {\bf Partial distance correlation in deep networks.} 

{
\section{Optimizing Distance Correlation in Neural Networks}
While distance correlation can be 
implemented in a differentiable way, 
and thereby used as an appropriate loss function in a neural network, we must take efficiency  
into account.
For two $p$ dimensional random variables, let the number of samples for the empirical estimate of DC be $n$. Observe that the total cost for computing $(a_{k,l})$ is $O(n^2 p)$, and the memory to store the intermediate matrices is also $O(n^2)$.
% Resnet \cite{he2016deep} corresponds to feature space dimensions higher than $d = 512$, while  Cifar-10 \cite{krizhevsky2009learning} dataset contains $n=60000$ samples. This makes it infeasible to compute the empirical estimate of distance correlation: it requires more than $1.8e^{12}$ multiplications and $13.4 GB$ to store the intermediate matrices. 
So, we use a stochastic estimate of DC by averaging over minibatches, with each minibatch containing $m$ samples. We describe why this approximation is sensible. 

\noindent{\bf Notation. }
We use $\Theta_X,\Theta_Y$ to denote the parameters of the neural networks, and $X,Y$ as features extracted by the respective neural networks. Let the minibatch size be $m$, and the dataset $\mathcal{D}=(\mathcal{D_X},\mathcal{D_Y})$ be of size $n$. 
% Let, $X\in \mathbb{R}^{m\times p}, Y \in \mathbb{R}^{m\times q}$, with $p,q$ be the dimension of features. 
We use $(x_t,y_t)_{t=1}^T,x_t\subset \mathcal{D_X},y_t\subset \mathcal{D_Y}$ to represent the data samples at step $t$, $T$ is the total number of training steps. The distance matrices $A_t,B_t$ are computed when given $X_t,Y_t$ using \eqref{eq:DC_A}, which is of dimension $m\times m$ for each minibatch. Further, we use $(X_t)_k$ to represent the $k^{th}$ element in $X_t$. And $(A_t)_{k,l}$ is the  $k^{th}$ row and $l^{th}$ column element in the matrix $A_t$.
% The trace of a square matrix is $\text{trace}(A_t)$, and the standard inner-product between two matrices $A_t,B_t$ is $\langle A_t, B_t \rangle = \text{trace}(A_t^T B_t)$.
The inner-product between two matrices $A,B$ is defined as $\langle A,B \rangle = \sum_{i,j}^{m} (A)_{i,j}(B)_{i,j}$.

\noindent{\bf Objective function.}
Consider the case where we minimize DC between 
two networks $\Theta_X,\Theta_Y$. Since the parameters between $\Theta_X,\Theta_Y$ are separable, we can use the block stochastic gradient iteration in \cite{xu2015block} with some simple modifications.

To minimize the distance correlation, we need to solve the following problem
\vspace{-1.5em}
\begin{align}
    &\min_{\Theta_X,\Theta_Y} \frac{\langle A(\Theta_X;x),B(\Theta_Y;y) \rangle}{\sqrt{\langle A(\Theta_X;x),A(\Theta_X;x)\rangle \langle B(\Theta_Y;y),B(\Theta_Y;y)\rangle}} \\
    (A)_{k,l} = & || (X)_k - (X)_l ||_2,\: X = \Theta_X(x), (B)_{k,l} = || (Y)_k - (Y)_l ||_2, \:\: Y = \Theta_Y(y) \nonumber
\end{align}
\vspace{-1.5em}

\noindent We slightly abuse the notation of $\Theta_X(x)$ as applying the network $\Theta_X$ onto data $x$, and reuse $A$ to simplify the notation $A(\Theta_X;x)$ and the distance matrix.
We can rewrite the expression (with $A$, $B$ defined above) using:
\vspace{-0.5em}
\begin{align}
    \min_{\Theta_X,\Theta_Y} \langle A, B \rangle \:\: \text{s.t.} \:\: & \max_{x\subset \mathcal{D_X}}\langle A,A\rangle\leq m; \:\: \max_{y\subset \mathcal{D_Y}}\langle B,B\rangle \leq m 
    % & \max_{y_t}\|B\|_{*}\leq m; \:\: \max_{y_t}\|B\|_2 \leq \beta/\sqrt{m}
\end{align}
\vspace{-1.5em}

\noindent where $(x,y)$ are the minibatch of samples from the data space $(\mathcal{D_X},\mathcal{D_Y})$.

% In the block stochastic gradient setup, we rewrite the following equation,
We can  rewrite the above into the following equation similar to (1) in \cite{xu2015block}.
\vspace{-0.5em}
\begin{align}
    \min_{\Theta_X,\Theta_Y}\Phi(\Theta_X,\Theta_Y) = \mathbb{E}_{x,y}f(\Theta_X,\Theta_Y;x,y) + \gamma(\Theta_X) + \gamma(\Theta_Y)
\end{align}
\vspace{-1em}

\noindent where $f(\Theta_X,\Theta_Y;x,y)$ is $\langle A, B \rangle$ and $\gamma(\Theta_X)$ encodes the convex constraint of network $\Theta_X$: $\max_{x\subset\mathcal{D_X}}\langle A,A\rangle\leq m$. Similarly, $\gamma(\Theta_Y)$ encodes $\max_{y\subset \mathcal{D_Y}}\langle B,B\rangle \leq m$. $\Phi(\Theta_X,\Theta_Y)$ is the constrained objective function to be optimized. 
% Throughout the paper, we let 
% \begin{align}
%     F(\Theta_X,\Theta_Y) = \mathbb{E}_{x,y}f(\Theta_X,\Theta_Y;x,y),\:\: \Gamma(\Theta_X,\Theta_Y)=\gamma(\Theta_X)+\gamma(\Theta_Y)
% \end{align}

\noindent{\bf Block stochastic gradient iteration.}
We adjust Alg. 1 from \cite{xu2015block} to our case in Alg. \ref{alg:bsg}. Since we will need the entire minibatch $(x_t,y_t)$ to compute the objective function, there will be no mean term when computing the sample gradient $\tilde {\mathbf{g}}_{X}^{t}$. Further, since both blocks $(\Theta_X,\Theta_Y)$ are constrained, line $3,5$ will use (5) from \cite{xu2015block}. The detailed algorithm is presented in Alg. \ref{alg:bsg}. 

% \vspace{-1.5em}
\renewcommand{\algorithmicrequire}{\textbf{Input:}}
\renewcommand{\algorithmicensure}{\textbf{Output:}}
\begin{algorithm*}
\caption{Block Stochastic Gradient for Updating Distance Correlation}\label{alg:bsg}
    \begin{algorithmic}[1]
        \REQUIRE Two neural network with starting point $\Theta_X^1,\Theta_Y^1$. Training data $\{(x_t,y_t)\}_{t=1}^T$, step size $\eta_X,\eta_Y$, and batch size $m$.
        \ENSURE $\tilde \Theta_X^T,\tilde \Theta_Y^T$
        \FOR {$t=1,\cdots,T$}
            \STATE Compute sample gradient for $\Theta_X$ \\
             \:\: $\tilde{\mathbf{g}}_{X}^{t} = \nabla_{\Theta_X} f(\Theta_X^t,\Theta_Y^t;x_t,y_t) $
            \STATE $\Theta_X^{t+1}=\argmin_{\Theta_X} \langle \tilde{\mathbf{g}}_{X}^{t} + \tilde{\nabla} \gamma_{X}(\Theta_X^t), \Theta_X-\Theta_X^t \rangle + \frac{1}{2\eta_X}\| \Theta_X - \Theta_X^t \|^2 $
            \STATE Compute sample gradient for $\Theta_Y$ \\
             \:\: $\tilde{\mathbf{g}}_{Y}^{t} = \nabla_{\Theta_Y} f(\Theta_X^{t+1},\Theta_Y^t;x_t,y_t) $
            \STATE $\Theta_Y^{t+1}=\argmin_{\Theta_Y} \langle \tilde{\mathbf{g}}_{Y}^{t} + \tilde{\nabla} \gamma_{Y}(\Theta_Y^t), \Theta_Y-\Theta_Y^t \rangle + \frac{1}{2\eta_Y}\| \Theta_Y - \Theta_Y^t \|^2 $
            % \STATE $X_{t+1} \leftarrow X_t + \alpha \eta {d_X}_t$ \COMMENT{\% $\alpha$ is chosen as $\min_{t,k,l} ({\|\frac{dA}{d{X_t}_k} - \frac{dA}{d{X_t}_l}\|})^{-1}$ {\color{red}{We need $\|A(X_t+\alpha \eta \partial_t \frac{dA}{dX})\| \leq \|A(X_t)+\eta \partial_t\| $}}}
        \ENDFOR
    \STATE $\tilde \Theta_X^T= \frac{1}{T} \sum_{t=1}^{T}\Theta_X^t$
    \STATE $\tilde \Theta_Y^T= \frac{1}{T} \sum_{t=1}^{T}\Theta_Y^t$
    \end{algorithmic}
\end{algorithm*}

\begin{proposition}
    After $T$ iterations of Algorithm \ref{alg:bsg} with step size $\eta_X=\eta_Y =\frac{\eta}{\sqrt{T}} < \frac{1}{L}$, for some positive constant $\eta<\frac{1}{L}$, where $L$ is the Lipschitz constant of the partial gradient of $f$, by Theorem. 6 in \cite{xu2015block}, we know there exists an index subsequence $\mathcal{T}$ such that:
    \vspace{-0.5em}
    \begin{align}
        \lim_{t\rightarrow \infty,t\in \mathcal{T}} \mathbb{E}[\text{dist}(\mathbf{0}, \nabla\Phi(\Theta_X^t,\Theta_Y^t))]=0
    \end{align}
    \vspace{-1em}
    
    \noindent where $\text{dist}(\mathbf{y},\mathcal{X})=\min_{\mathbf{x}\in \mathcal{X}}\| \mathbf{x}-\mathbf{y}\|$. 
    
    % Further, if $\mathbb{E}_{x,y}f(\Theta_X,\Theta_Y;x,y)$ is convex, by Theorem. 1 in \cite{xu2015block}, the following statement holds:
    % \vspace{-0.5em}
    % {\small
    % \begin{align}
    %     \mathbb{E}[\Phi(\tilde\Theta_X^T,\tilde\Theta_Y^T) - \Phi(\Theta_X^*,\Theta_Y^*)] &\leq D\eta \frac{1+\log T}{\sqrt{1+T}} + \frac{\| \Theta_X -\Theta_X^1 \|^2 + \| \Theta_Y -\Theta_Y^1 \|^2}{2\eta \sqrt{1+T}}
    % \end{align}
    % }%
    % \vspace{-1em}
    
    % \noindent where $\tilde \Theta_X,\tilde \Theta_Y$ is computed in Algorithm \ref{alg:bsg}, $\Theta_X^*,\Theta_Y^*$ are the optimum of the desired function, and $D$ is a constant depending on $\|( \Theta_X^*;\Theta_Y^*) \|$
    % where $A_*$ is the optimum of the desired function and $\tilde A$ is defined in the Algorithm \ref{alg:msg}. 
\end{proposition}    

But empirically, we find that simply applying Stochastic Gradient Decent (SGD) is sufficient, but this choice is 
available to the user.

}
\section{Independent Features Help Robustness}
\label{sec:diverge}

{\bf Goal.} We show how distance correlation can help us train multiple deep networks that learn {\bf mutually independent} features, roughly similar to  
finding diverse $M$-best solutions in 
structured SVM models \cite{martin2016learning}. 
We describe how such an approach can lead to better robustness against adversarial attacks.

\noindent {\bf Rationale.} Recently, several efforts have explored generating 
of adversarial examples that can transfer to different networks and how to defend against 
such attacks \cite{demontis2019adversarial,shumailov2019sitatapatra,chan2020thinks}. It is often observed that an adversarial sample for 
{one trained network is relatively easy to transfer to another network with the same architecture \cite{demontis2019adversarial}}. Here, we show that even for as few as two networks (same architecture; trained on the same data), we can, to some extent, prevent adversarial examples from transferring between them by seeking independent features. 

\noindent{\bf Setup.} We formulate the problem considering a classification task as an example. Given two deep neural networks with the same architecture denoted as $f_1(\cdot), f_2(\cdot)$, we train them using image-label pairs $(x, y)$ using the cross-entropy loss $\text{Loss}_{\text{CE}}$.
% \begin{equation}
%     \text{Loss}_{\text{CE}} = \text{Cross-Entropy}(f(x), y)
%     \label{ce_loss}
% \end{equation}
If we train $f_1$ and $f_2$ using only the cross-entropy loss, the adversarial examples generated on $f_1$ can relatively easily transfer to $f_2$ (see the performance of ``Baseline'' in Table \ref{tab:adv_transfer}).
To enforce $f_1$ and $f_2$ to learn independent features, let the extracted feature of $x$ in some intermediate layer of $f$ be given as $g(x)$ (in this section we use the feature before the last fully connected layer as an example). We can still train $f_1$ using $\text{Loss}_{\text{CE}}$, and then, we train $f_2$ using,
\vspace{-0.5em}
\begin{equation}
    \text{Loss}_{\text{total}} = \text{Loss}_{\text{CE}}(f_2(x),y) + \alpha \cdot \text{Loss}_{\text{DC}}(g_1(x), g_2(x))
    \label{ce_dc_loss}
\end{equation}

\vspace{-1em}
\noindent where $\alpha$ is a constant scalar and Loss$_{\text{DC}}$ is the distance correlation from Def. \ref{def:1}. 
%After such training, $f_2$ will learn independent features from $f_1$ and the results in Table XXX show that the adversarial samples generated on $f_1$ has worse transferability to $f_2$.
Note that we do not require $g_1(x)$ and $g_2(x)$ to be in the same dimension, so in principle we could 
easily use features from different layers for these two networks.

\noindent\textbf{Experimental settings.} We first conduct experiments on CIFAR10 \cite{krizhevsky2009learning} using Resnet 18 \cite{he2016deep}. We then use four different architectures (mobilenet-v3-small \cite{howard2019searching}, efficientnet-B0 \cite{tan2019efficientnet}, Resnet 34, and Resnet152) and train them on ImageNet \cite{krizhevsky2012imagenet}. For each network architecture, we first train two networks using only $\text{Loss}_{\text{CE}}$. Next, we train a network using only $\text{Loss}_{\text{CE}}$ before  training a second network using the loss in \eqref{ce_dc_loss}. On CIFAR10, we utilize
the SGD optimizer with momentum $0.9$ and train for $200$ epochs using an initial learning rate $0.1$ with a cosine learning rate scheduler \cite{paszke2019pytorch}. The mini-batch size is set to $128$. On ImageNet \cite{krizhevsky2012imagenet}, we train for $40$ epochs using an initial learning rate $0.1$, which decays by $0.1$ every 10 epochs. The mini-batch size is $512$. Our $\alpha$ in \eqref{ce_dc_loss} is set to $0.05$ for all cases. For each combination of the dataset and the network architecture, we train two networks $f_1$ and $f_2$, after which we generate adversarial examples on $f_1$ and use them to attack $f_2$ and measure its classification accuracy. We construct a baseline by training $f_1$ and ${f_2}_{\rm Baseline}$ without constraints. And train ${f_2}_{\rm Our}$ using \eqref{ce_dc_loss} to learn independent features w.r.t. $f_1$. We report performance under two widely used attack methods: fast gradient sign method (FGM) \cite{goodfellow2014explaining} and projected gradient descent method (PGD) \cite{madry2017towards}, where the latter is 
considered among the strongest attacks. The scale $\epsilon$ of the adversarial perturbation is chosen from $\{0.03,0.05,0.1\}$ and the maximum number of iterations of PGD is set to $40$. 

\noindent{\textbf{Results.} The results are shown in Table \ref{tab:adv_transfer}. We see that we get significant improvement in accuracy over the baseline under adversarial attacks, with comparable performance on clean inputs. Notably, our method achieves more than $10\%$ absolute improvement in accuracy under PGD attack on Resnet-18 and Mobilenet-v3-small. This provides evidence supporting the benefits of  enforcing the networks to learn independent features using our distance correlation loss.}

\begin{table*}[t]
\caption {\footnotesize The  test accuracy (\%) of a model $f_2$ on the adversarial examples generated using $f_1$ with  the same architecture. ``Baseline'': train without constraint. ``Ours'': $f_2$ is independent to $f_1$. ``Clean'': test accuracy without adversarial examples.} \label{tab:adv_transfer} 
\centering
\scalebox{0.75}{
\begin{tabular}{ccc|c|cc|cc|cc}
\topline\myrowcolour
Dataset & Network & Method   &
Clean &
FGM$_{\epsilon=0.03}$ & PGD$_{\epsilon=0.03}$ & FGM$_{\epsilon=0.05}$ & PGD$_{\epsilon=0.05}$ & FGM$_{\epsilon=0.1}$   & PGD$_{\epsilon=0.1}$\\
CIFAR10 &  Resnet 18 &  Baseline  & 89.14     & 72.10    & 66.34    & 62.00       & 49.42       & 48.23 & 27.41\\
CIFAR10 &  Resnet 18 &  Ours  & 87.61     & \textbf{74.76}     & \textbf{72.85}    & \textbf{65.56}      & \textbf{59.33}       & \textbf{50.24} & \textbf{36.11} \\
% \topline\myrowcolour
% Dataset & Network & Method   &
% Clean &
% FGM_{\epsilon=0.003} & PGD_{\epsilon=0.003} & FGM_{\epsilon=0.005} & PGD_{\epsilon=0.005} & FGM_{\epsilon=0.01}   & PGD_{\epsilon=0.01}\\
\midtopline
ImageNet &  Mobilenet-v3-small &  Baseline  & 47.16     & 29.64     & 30.00    & 23.52       & 24.81      & 13.90  & 17.15 \\
ImageNet &  Mobilenet-v3-small &  Ours  & 42.34     & \textbf{34.47}     & \textbf{36.98}    & \textbf{29.53}       & \textbf{33.77}       & \textbf{19.53} & \textbf{28.04} \\
\midtopline
ImageNet & Efficientnet-B0 &  Baseline  & 57.85     & 26.72     & 28.22    & 18.96       & 19.45     & 12.04 & 11.17 \\
ImageNet & Efficientnet-B0 &  Ours  & 55.82     & \textbf{30.42}    & \textbf{35.99}    & \textbf{22.05}       & \textbf{27.56}       & \textbf{14.16} & \textbf{17.62} \\
\midtopline
ImageNet &  Resnet 34 &  Baseline  & 64.01     & 52.62     & 56.61    & 45.45       & 51.11       & 33.75 & 41.70\\
ImageNet &  Resnet 34 &  Ours  & 63.77     & \textbf{53.19}     & \textbf{57.18}   & \textbf{46.50}     & \textbf{52.28}       & \textbf{35.00} & \textbf{43.35} \\
\midtopline
ImageNet & Resnet 152 &  Baseline  & 66.88     & 56.56     & 59.19    & 50.61       & 53.49     & 40.50 & 44.49 \\
ImageNet & Resnet 152 &  Ours  & 68.04     & \textbf{58.34}    & \textbf{61.33}    & \textbf{52.59}       & \textbf{56.05}       & \textbf{42.61} & \textbf{47.17} \\
\bottomline 
\end{tabular}
% \label{tab:adv_transfer}
}
\vspace{-2.5em}
\end{table*}

\begin{figure}[!b]
\vspace{-1em}
        \centering
               \includegraphics[width=0.98\columnwidth]{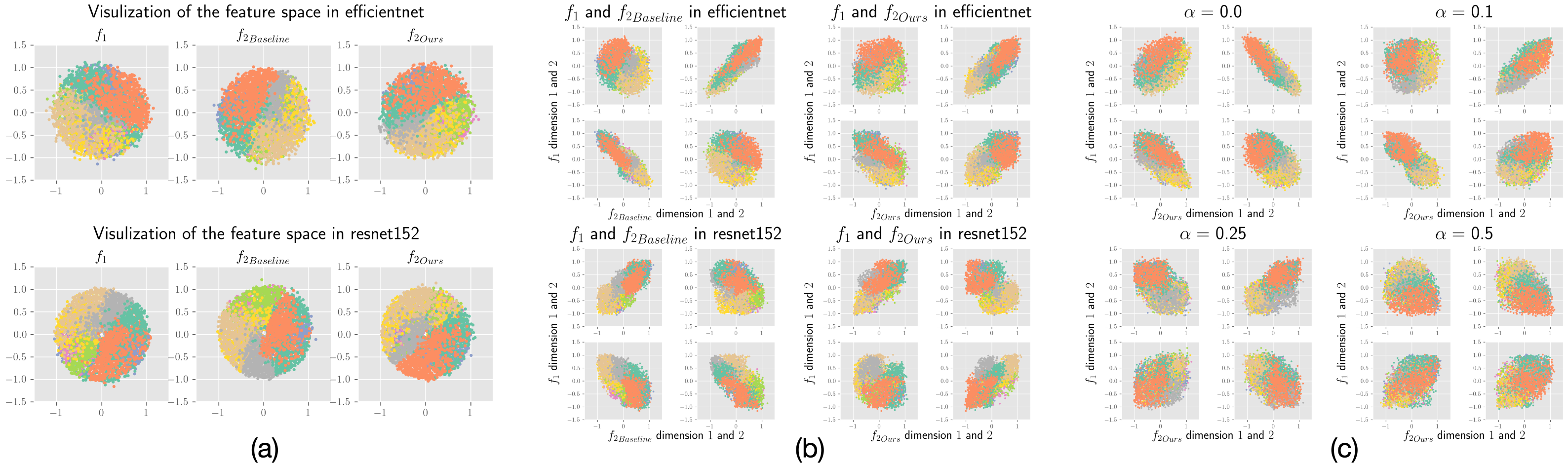}
\vspace{-1em}
               \caption{\footnotesize Picasso visualization of features space and the correlation between different models. {\bf{(a)}} Feature space distribution. {\bf{(b)}} Cross-correlation between the feature space of $f_1$ and $f_2$ trained with/without DC. We get better independence. {\bf{(c)}} By increasing the balance parameter $\alpha$ of DC loss, Mobilenet is more independent to $f_1$. }
               \label{fig:vis_diverge}
    \vspace{-1em}
\end{figure}

In Fig. \ref{fig:vis_diverge}, we show  correlation results using Picasso \cite{henderson2017picasso,chari2021specious} to lower the dimension of features for each network. The embedding dimension is $2$ for visualization. In Fig. \ref{fig:vis_diverge}(a), we show the embedding of different networks. $f_1$ represents the network to generate the adversarial examples. ${f_2}_{\rm Baseline}$ denotes the baseline network, trained without distance correlation constraint. Also, ${f_2}_{Ours}$ is the same network trained to be independent to $f_1$. In Fig. \ref{fig:vis_diverge}(b), we visualize the correlation between $f_1$ and ${f_2}_{Baseline}$ for each dimension, and the correlation between $f_1$ and ${f_2}_{Ours}$. If the scatter plot looks  circle-like, we can infer that the two models are independent. We see that in different networks, the use of DC shows stronger independence. From Fig. \ref{fig:vis_diverge}/Tab. \ref{tab:adv_transfer}, we also see that the more independent the models are, the better is the gain for transferred attack robustness.

%We also demonstrate that the ensemble of multiple models that learn independent features can be hard to attack. ?(to be determined by results)

% Outline: ensemble can act as a way to defend and it's orthogonal to the defense techniques proposed to train a single network. However, ensembling should be used on  networks which learn different features. e.g., if the networks are the same, transfer attack is easy, so ensembling doesn't help (essentially it's useful because adversarial examples don't transfer too well.) exp demonstrates that it transfer worse (also explains why different network hard to transfer?). Then show that by encorcing independent features, the ensembled network has large improvement on PGD.

% {\color{red}
% From here, are some examples about how we extend DC to different aspect of deep network. Some modification is need, and some benefit shows up. 
% }

% {\color{blue}
% Discuss about how to take into consideration of different networks such that each networks are independent. 
% Use adversarial methods to show the benefit of diverge training.
% }

\section{Informative Comparisons between Networks}
% {\color{red}
% First talk about the baseline model and their performance. Re-produce the Fig. 1 and Fig. 2. Then go one step further about take information of network $Y$ out of $X$, how to do this. Use Grad-CAM to show the difference and the value in the coming table. Discuss what that means.
% }

%As we mentioned in the Introduction section, people are curious about understanding the
{\bf Overview.} As discussed in \S\ref{sec:intro}, there 
is much 
interest in understanding whether two different 
models learn similar concepts from 
the data -- for example, 
whether vision Transformers ``see'' similar to convolutional neural networks \cite{raghu2021vision}. 
Here, we first follow \cite{raghu2021vision} and discuss similarities between different layers of ViT and ResNets 
using distance correlation. Next, we
investigate that after taking out the influence of Resnets from ViT (or vice versa), 
what are the residual learned concepts 
remaining in the network.

\subsection{Measure Similarity between Neural Networks}
\noindent{\bf Goal.} We first want to understand whether ViTs represent features across all layers differently from CNNs (such as Resnets). However, analyzing the features in the hidden layers can be challenging, because the features are spread across neurons. Also, different layers have different numbers of neurons. Recently, \cite{raghu2021vision} applied the Centered Kernel Alignment (CKA) for this task. CKA is effective because it involves no constraint on the number of neurons. It is also independent to the orthogonal transformations of representations. Here, we want to demonstrate that distance correlation is a reasonable alternative for CKA in these settings.

\begin{figure}[!b]
% \vspace{-1em}
        \centering
               \includegraphics[height=0.125\columnwidth]{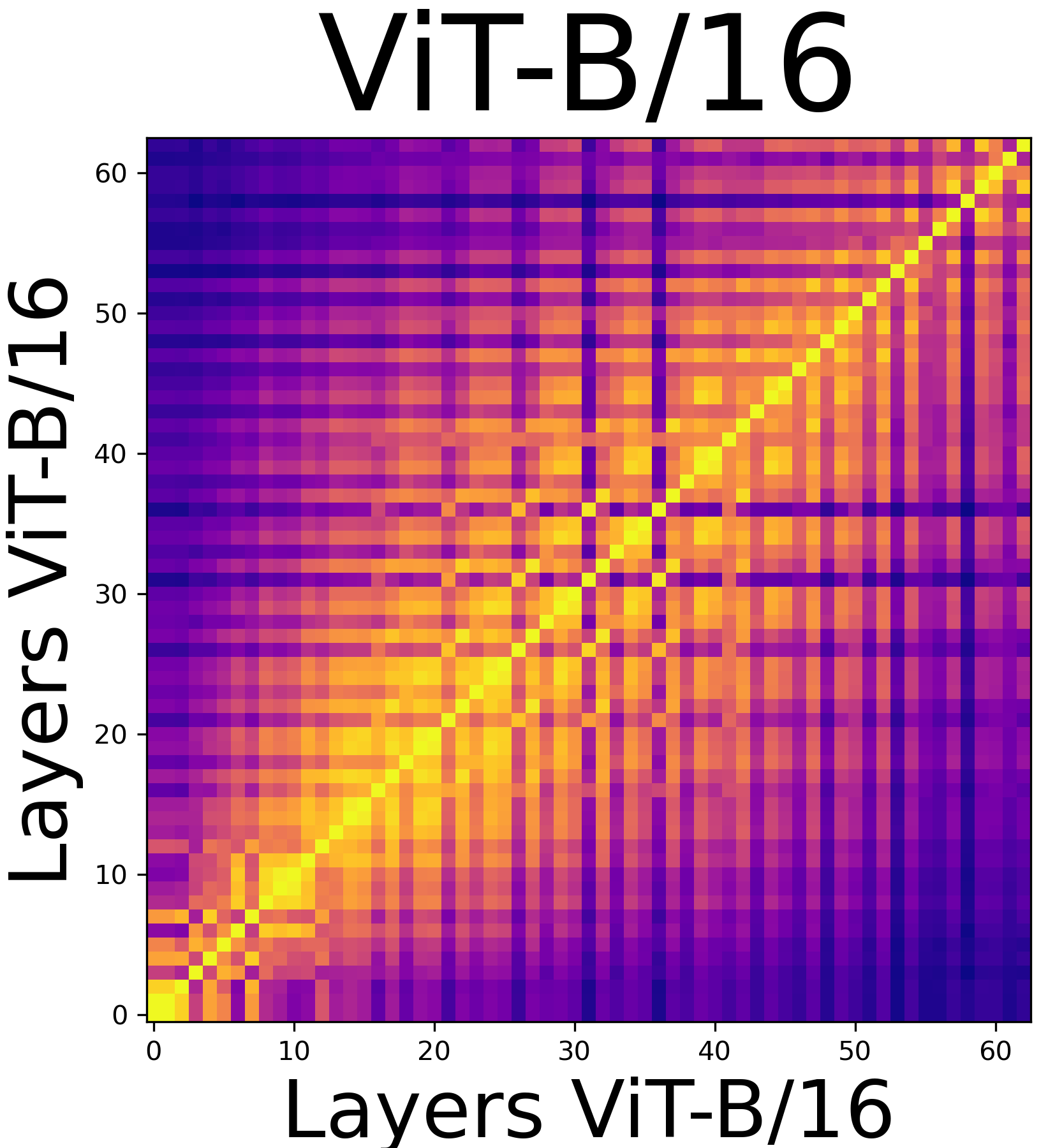}
               \includegraphics[height=0.125\columnwidth]{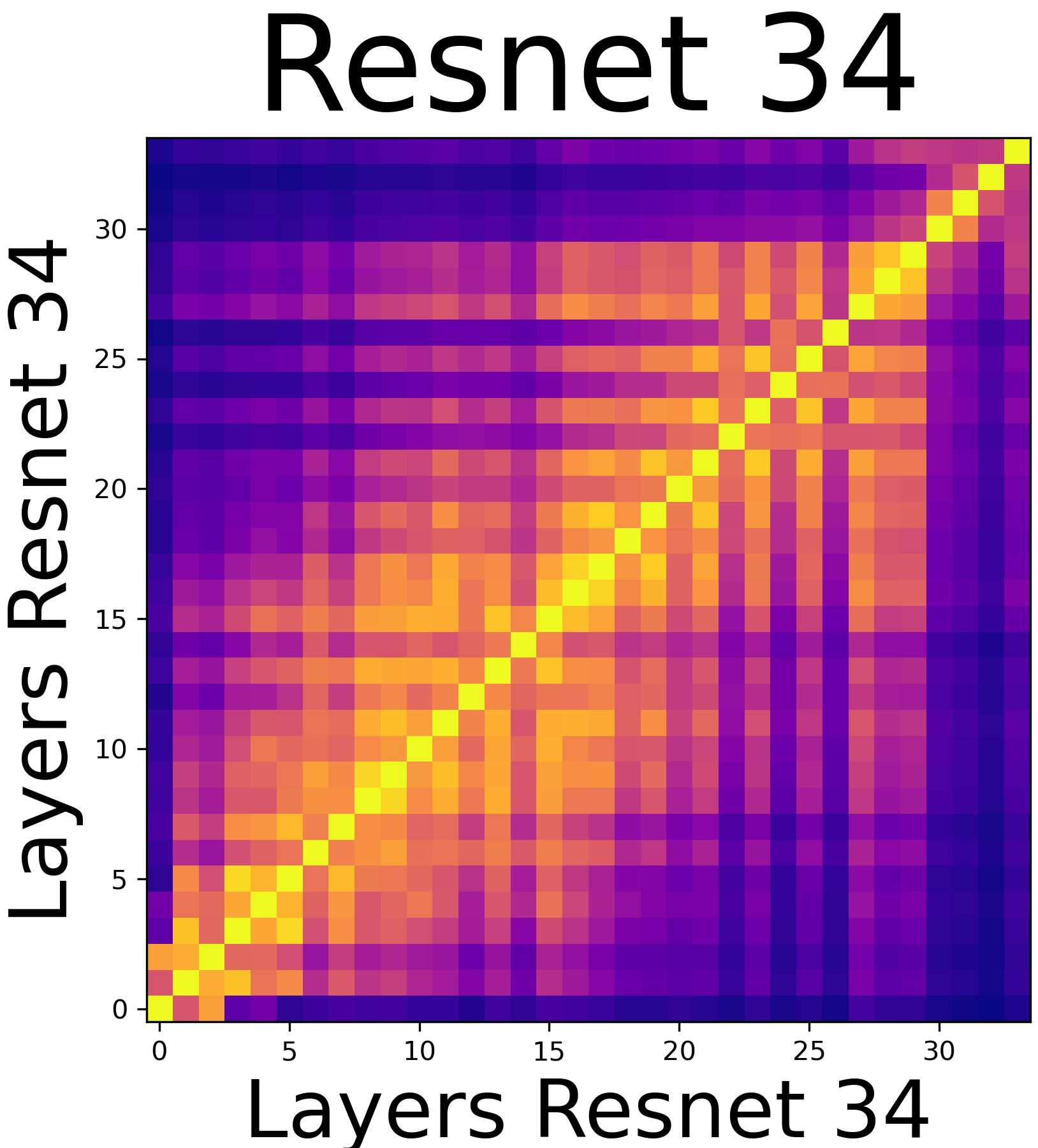}
               \includegraphics[height=0.125\columnwidth]{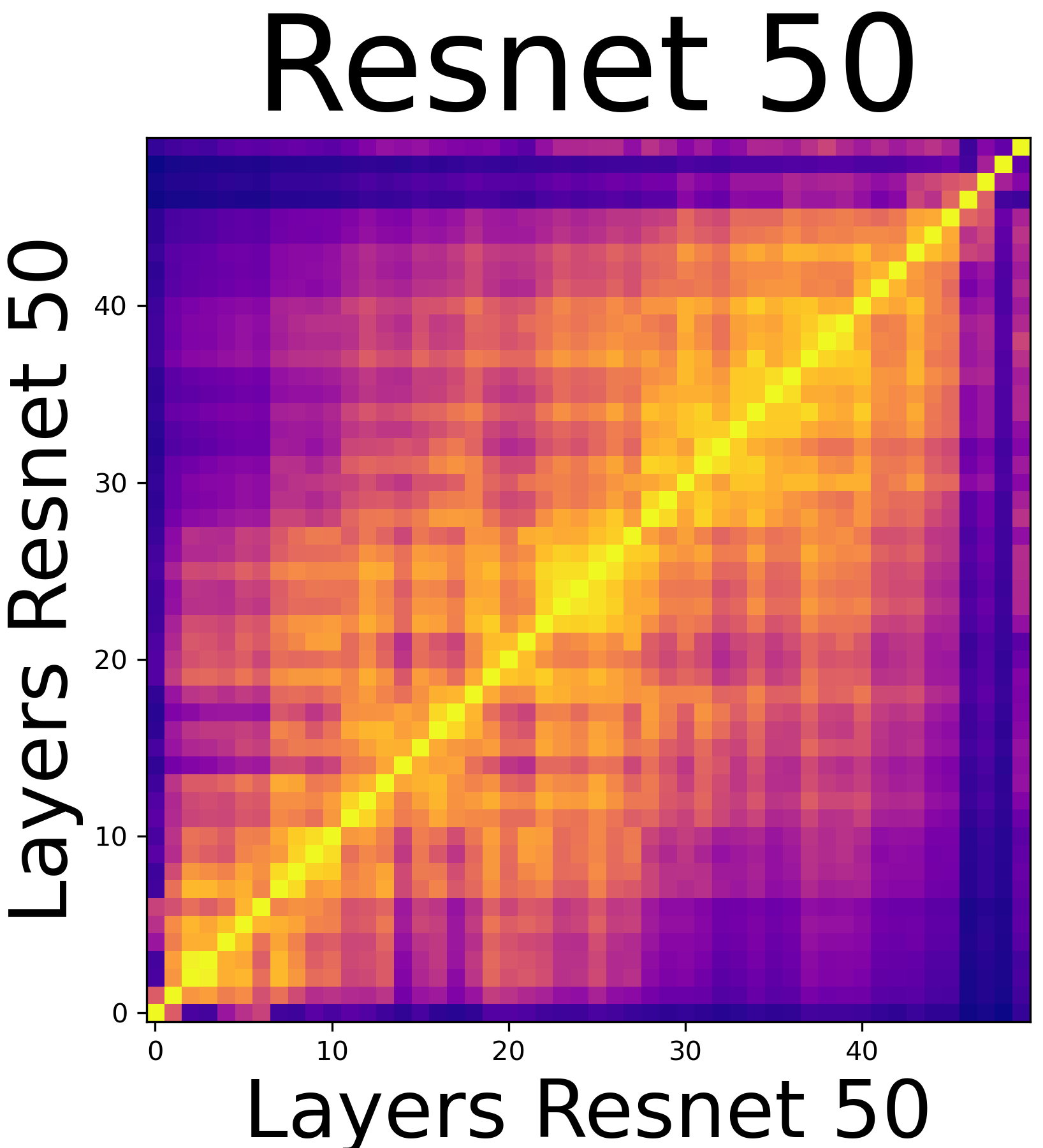}
               \includegraphics[height=0.125\columnwidth]{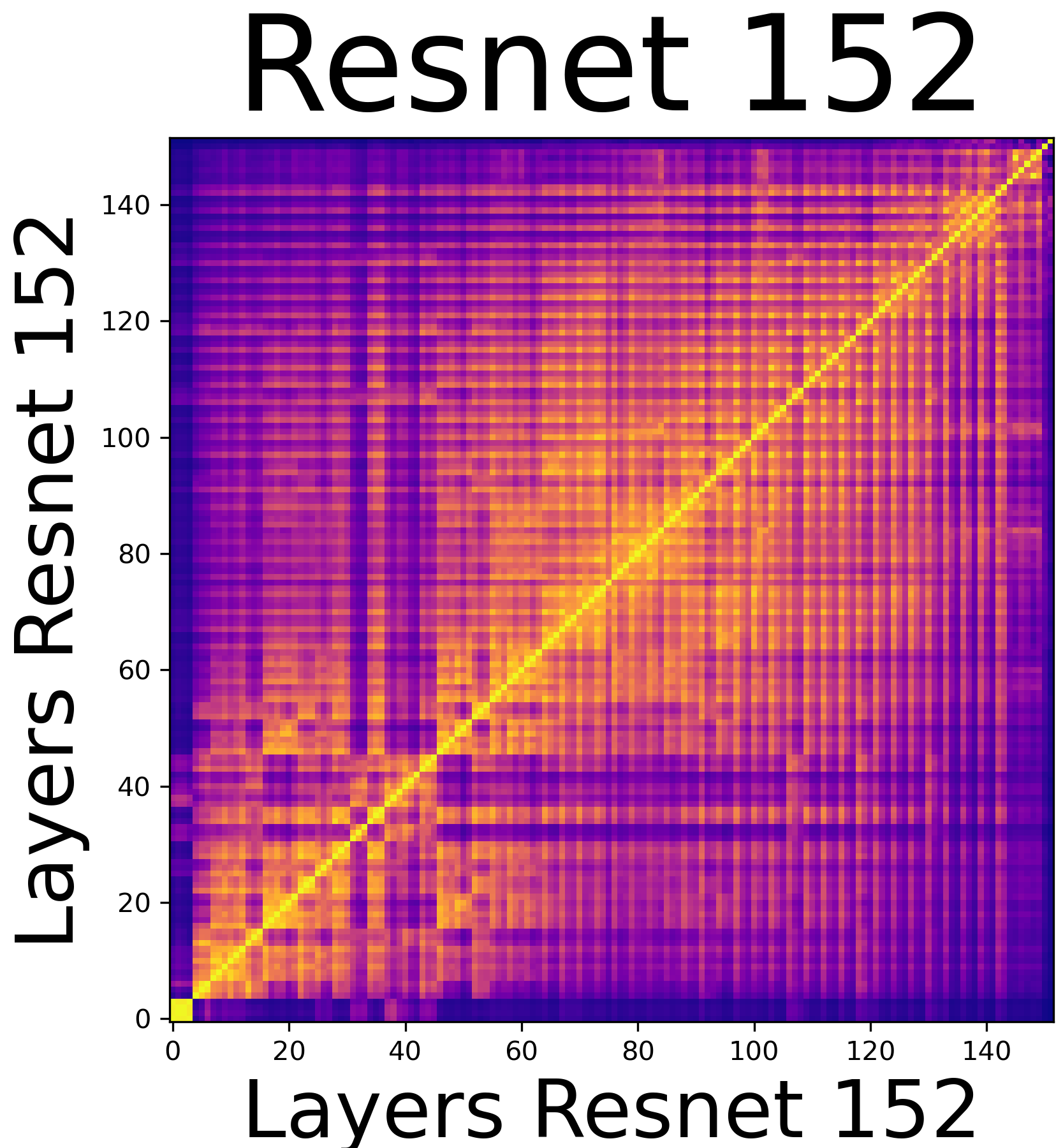}
               \includegraphics[height=0.02\columnwidth]{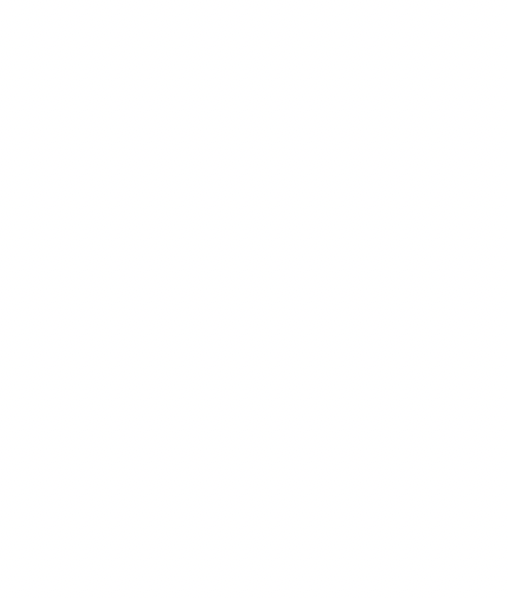}
               \includegraphics[height=0.13\columnwidth]{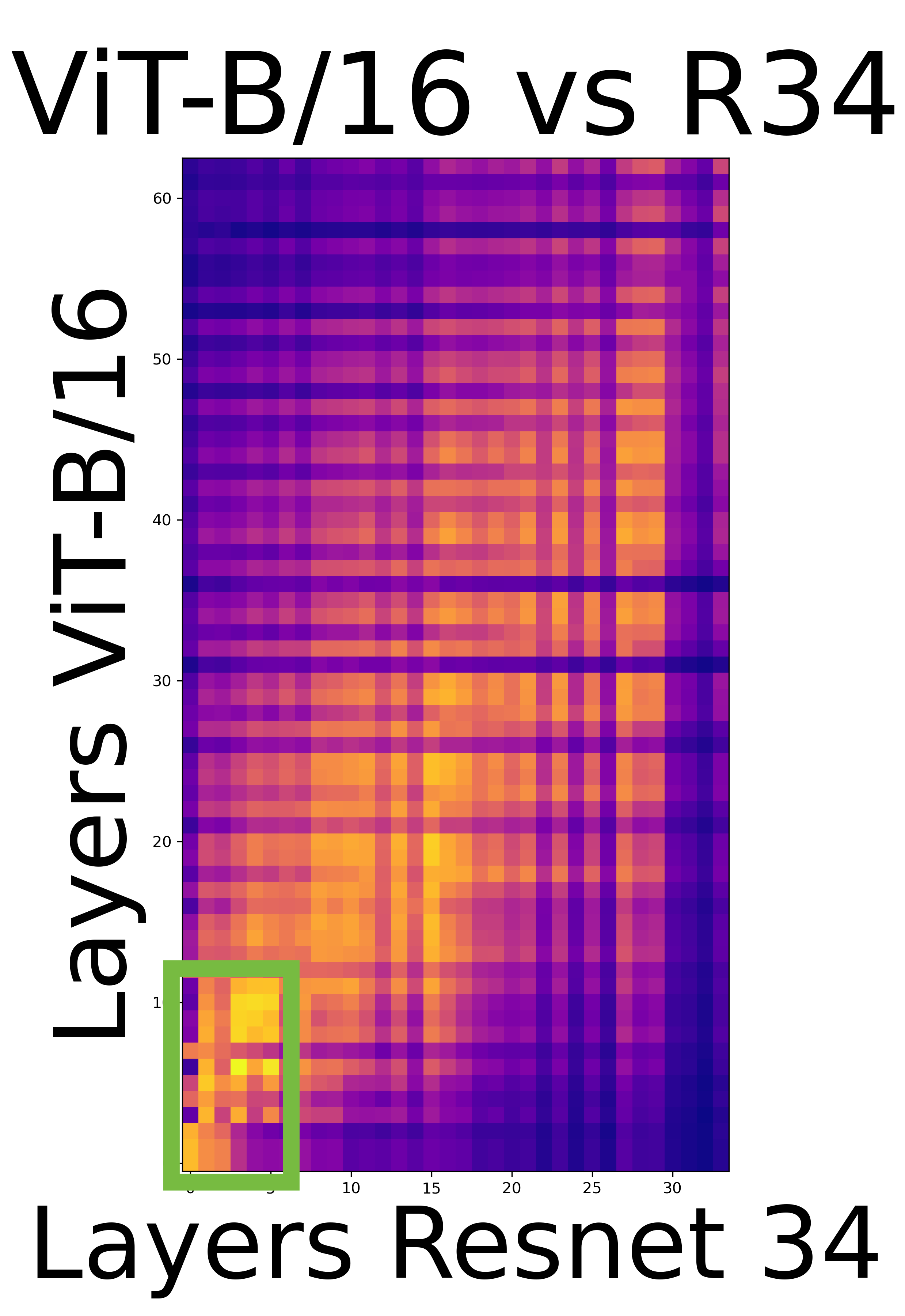}
               \includegraphics[height=0.13\columnwidth]{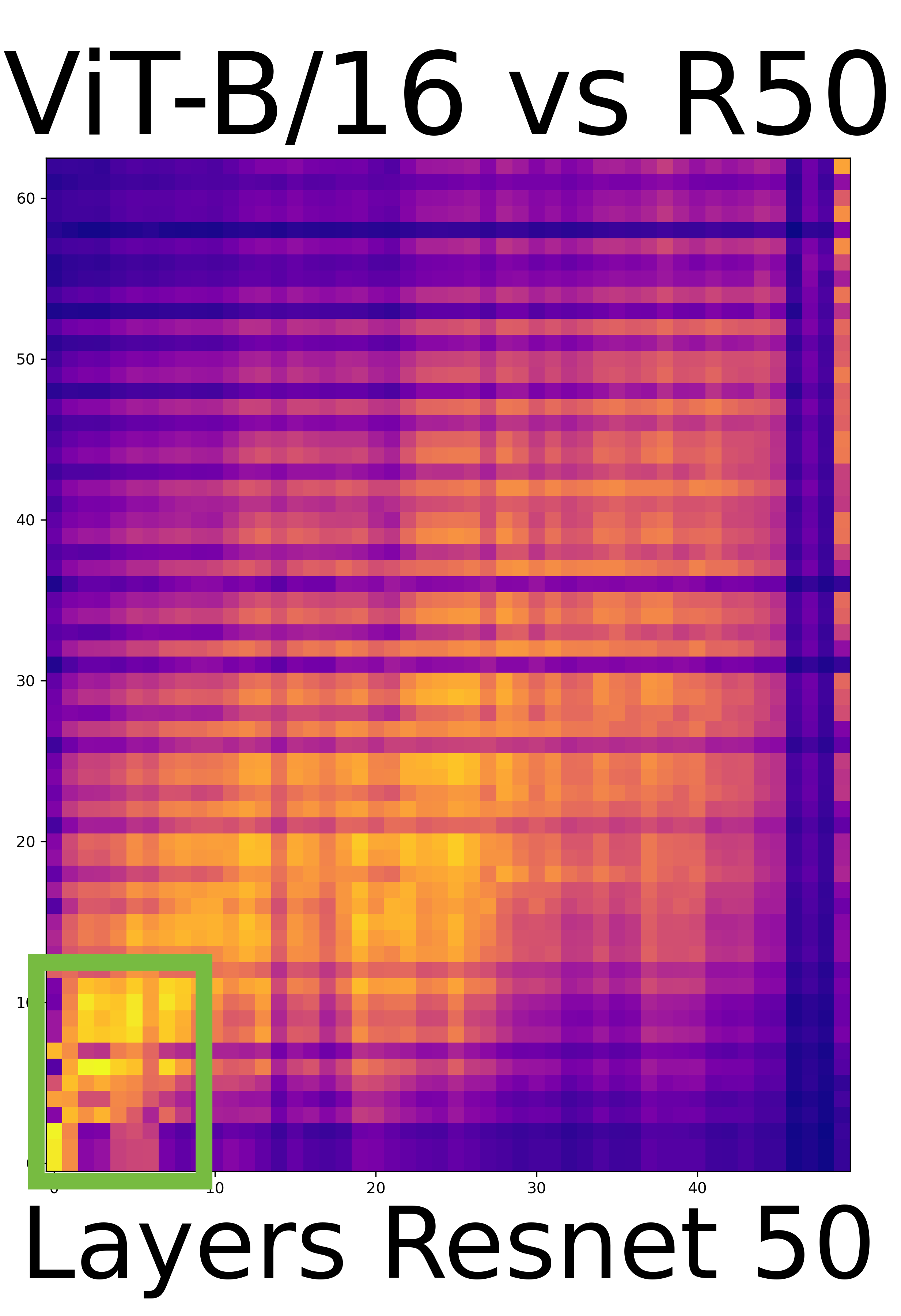}
               \includegraphics[height=0.13\columnwidth]{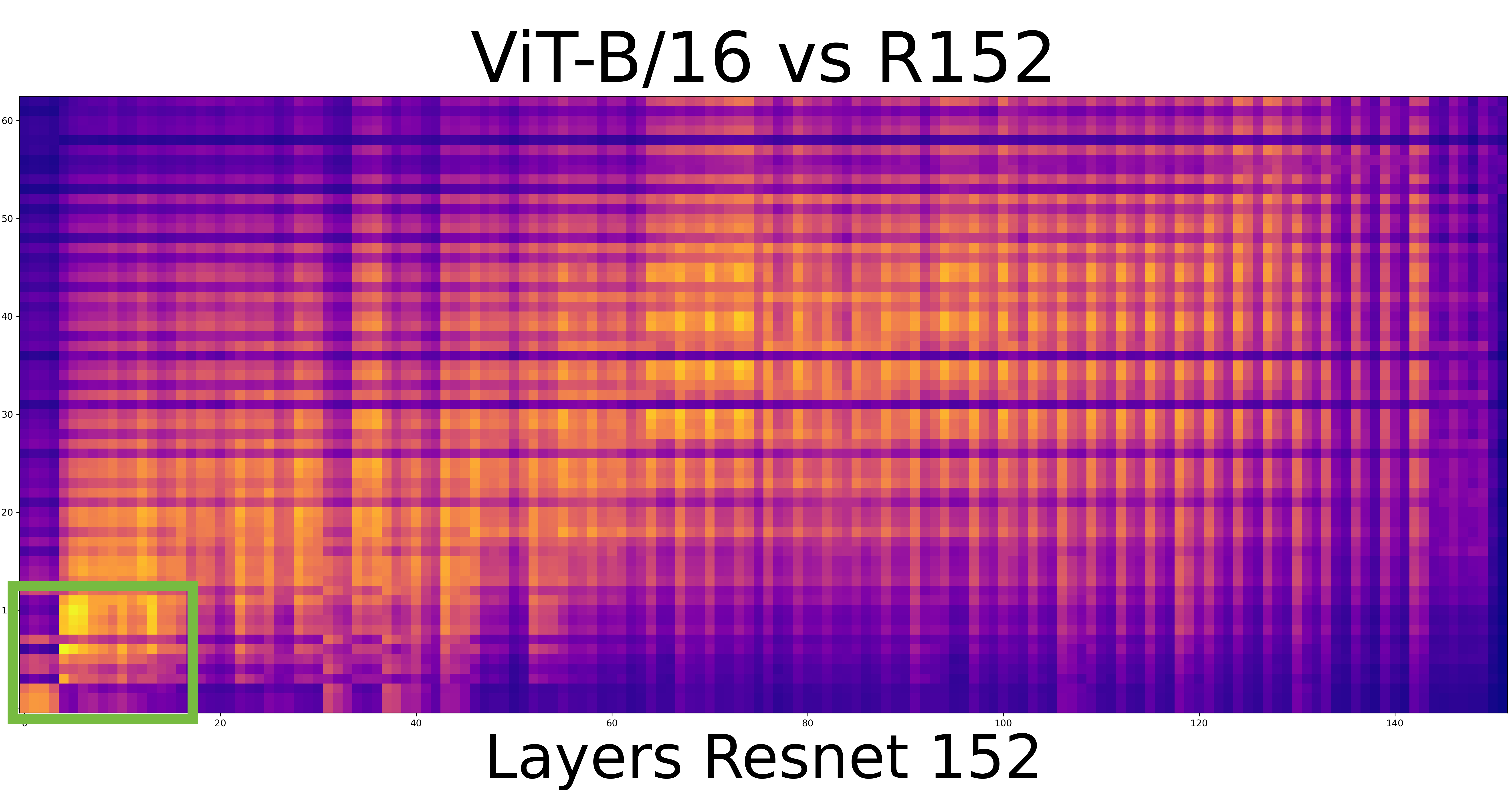}
               \includegraphics[height=0.12\columnwidth]{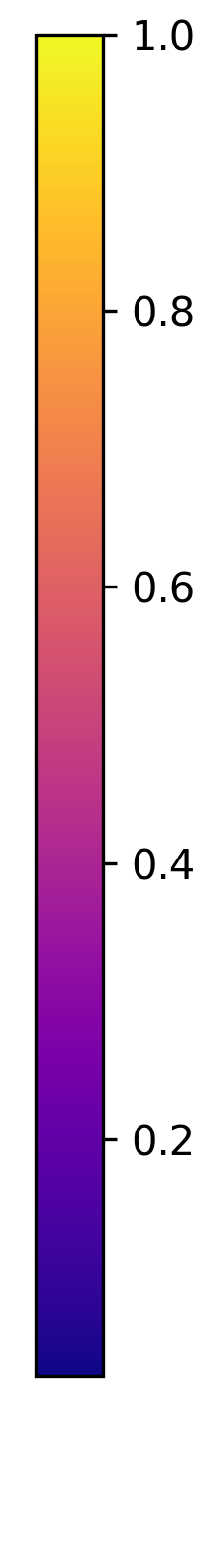}
\vspace{-1em}
               \caption{\footnotesize {\bf (a) Left 4:}  Similarity between layers within one single model. ViT can be split into small blocks and the similarity from shallow layers to the deeper layers is higher. Most Resnet models show few large blocks in the network, and the last few layers share minimal similarity with the shallow layers. {\bf (b) Right 3:} Similarity between layers across ViT and Resnets. In the initial $1/6$ layers (highlighted in green), the two networks share high similarity. And the last few layers share the least similarity}
               \label{fig:sim1}
    % \vspace{-1em}
\end{figure}

\noindent{\bf Experimental settings.} First, as described in \cite{raghu2021vision}, we show that similarity between layers within a single neural network can be assessed 
using distance correlation (see Fig. \ref{fig:sim1}(a) ). We pick ViT Base with patch 16, and three commonly used Resnets. All networks are pretrained on ImageNet. For ViT, we pick the embedding layer and all the normalization, attention, and fully connected layers within each block. The total number of layers is $63$. For Resnets, we use all convolutional layers and the last fully connected layer, which is the same counting method to build Resnet models.

\noindent{\bf Results (a).} Our findings add to those from \cite{raghu2021vision}. Using distance correlation, we find that the ViT layers can be split into small blocks and the similarity between different blocks from shallow layers to the deeper layers is higher. For most Resnets, the feature similarity shows that there are a few large blocks in the network, which contains more than 30 layers each, and the last few layers share minimal similarity with the shallow layers. 

\noindent{\bf Results (b).} After within-model distance correlation, we perform across-model distance correlation comparisons between ViT and Resnets, see Fig. \ref{fig:sim1}(b). We notice that in the initial $1/6$ layers, the two networks share high similarities. But later, the similarity spreads across all different layers between ViT and Resnets. Notably, the last few layers share the least similarity between two networks.

By using the distance correlation to calculate the heatmap of the similarity matrices, we can qualitatively describe the difference between the patterns of the features in different layers from different networks. What is even more interesting is to quantitatively show the difference, for example, to answer which network contains more information for the ground truth classes. We discuss this next.

% \begin{figure}[!hbt]
%         \centering
%               \includegraphics[height=0.246\columnwidth]{Figures/ViT_Resnet34.png}
%               \includegraphics[height=0.246\columnwidth]{Figures/ViT_Resnet50.png}
%               \includegraphics[height=0.246\columnwidth]{Figures/ViT_Resnet152.png}
%               \caption{\footnotesize  Similarity between layers across ViT and Resnets. In the initial $1/6$ layers, the two networks share high similarity. In the following layers, the similarity spreads across all different layers. And the last few layers share the least similarity}
%               \label{fig:sim2}
% \end{figure}

\subsection{What Remains When ``Taking out'' $Y$ from $X$}

\noindent{\bf Goal.} Even measuring information contained in one neural network is challenging, 
and often tackled by measuring the accuracy on the test dataset. But the association between accuracy and the information contained in a network may be weak. 
Based on existing literature, conditioning 
one network w.r.t. another remains unresolved. 
Despite the above challenges, we can 
indeed measure the similarity between the features of the network $X$ and the ground truth labels. If the similarity is higher, we can say that the feature space of $X$ contains more information regarding the true labels. Distance correlation enables this. 
Interestingly, partial 
distance correlation extends this idea to 
multiple networks allowing us to 
approach the ``conditioning'' question posed above. 

\noindent{\bf Rationale/setup.} 
Here, we choose the last layer before the final fully-connected layer as the feature layer similar to the setup in \S \ref{sec:diverge}. 
% The main reason is that earlier layers might learn only local information without taking the whole image into consideration. 
Our first attempt involved directly applying the distance correlation measurement to feature $X$ and the one-hot ground truth embedding. However, the one-hot embedding for the label contains very little information, e.g., it does not show the difference between ``cat'' vs. ``dog'' and ``cat'' vs. ``airplane''. So, we use the pretrained BERT \cite{devlin2018bert} to linguistically embed the class labels into the hidden space. We then measure the distance correlation between the feature space of $X$ and the pretrained hidden space $GT$. $\mathcal{R}^2(X,GT) = \frac{m}{n} \sum_{t=1}^{n/m} dCor(x_t,gt_t)$ where $x_t$ is the feature for one minibatch, and $gt_t$ is the BERT embedding vector of the corresponding label.  
To further extend this metric to measure the ``remaining'' or residual information, we apply the partial distance correlation calculation by 
% projecting the feature space $X$ to $Y$ in the Hilbert space of the $\mathcal{U}$-centered matrix $\tilde A$ and $\tilde B$. 
removing $Y$ out of $X$, or say $X$ conditioned on $Y$. 
Then, we have 
$\mathcal{R}^2\left( (X|Y),GT\right) = \frac{m}{n} \sum_{t=1}^{n/m} dCor\left( (x_t|y_t),gt_t\right)$
% $= \frac{m}{n} \sum_{t=1}^{n/m} (P_{y_t^{\perp}}(x_t) \cdot \tilde{gt}_t )/ (\|P_{y_t^{\perp}}(x_t)\|  \|\tilde{gt}_t\|)$
% , where $P_{y^{\perp}}(x)$ represents the $\mathcal{U}$-centered matrix for the residual features, and $\tilde{gt}$ is the $\mathcal{U}$-centered matrix for the ground truth embedding vectors. This capability has not been shown before.
using \eqref{eq:pdcov}. This capability has  not been shown before.

\begin{table}[!t]
\vspace{-1em}
\caption {\footnotesize Partial DC between the network $\Theta_X$ conditioned on the network $\Theta_Y$, and the ImageNet class name embedding. The higher value indicates the more information.} \label{tab:pdc} 
\begin{adjustbox}{max width=1.0\columnwidth}
\begin{threeparttable}
\centering
\scalebox{0.85}{
\setlength\tabcolsep{1.5pt}
\begin{tabular}{cc|cc|cc}
\topline\myrowcolour
Network $\Theta_X$ & Network $\Theta_Y$ &{\scriptsize $\mathcal{R}^2(X, GT)$}&{\scriptsize $\mathcal{R}^2(Y, GT)$}&{\scriptsize $\mathcal{R}^2((X|Y), GT)$}&{\scriptsize $\mathcal{R}^2((Y|X), GT)$}\\
ViT$^1$     & Resnet 18$^2$   & 0.042     & 0.025    & 0.035       & 0.007 \\
ViT         & Resnet 50$^3$   & 0.043     & 0.036    & 0.028       & 0.017 \\
ViT         & Resnet 152$^4$  & 0.044     & 0.020    & 0.040       & 0.009 \\
\midtopline
ViT         & VGG 19 BN$^5$  & 0.042     & 0.037    & 0.026       & 0.015 \\
ViT         & Densenet121$^6$ & 0.043     & 0.026    & 0.035       & 0.007 \\
\midtopline
ViT large$^7$   & Resnet 18   & 0.046     & 0.027    & 0.038       & 0.007 \\
ViT large   & Resnet 50   & 0.046     & 0.037    & 0.031       & 0.016 \\
ViT large   & Resnet 152  & 0.046     & 0.021    & 0.042       & 0.010 \\
ViT large   & ViT         & 0.045     & 0.043    & 0.019       & 0.013 \\
\midtopline
ViT+Resnet 50$^8$ & Resnet 18  & 0.044     & 0.024    & 0.037       & 0.005 \\
\midtopline
Resnet 152  & Resnet 18   & 0.019     & 0.025    & 0.013       & 0.020 \\
Resnet 152  & Resnet 50   & 0.021     & 0.037    & 0.003       & 0.030 \\
Resnet 50   & Resnet 18   & 0.036     & 0.025    & 0.027       & 0.008 \\
Resnet 50   & VGG 19 BN   & 0.036     & 0.036    & 0.020       & 0.019 \\
\bottomline 
\end{tabular}
}
\begin{tablenotes}
      \footnotesize
      \item \qquad Accuracy: \quad 1. 84.40\%; 2. 69.76\%; 3. 79.02\%; 4. 82.54\%;
      \item \qquad \hskip5.7em 5. 74.22\%; 6. 75.57\%; 7. 85.68\%; 8. 84.13\% 
\end{tablenotes}
\end{threeparttable}
\end{adjustbox}
\vspace{-2em}
\end{table}

\noindent{\bf Experimental settings.} 
In order to measure the information remaining when conditioning network $\Theta_Y$ out of $\Theta_X$, we first use pretrained networks on ImageNet. We use the validation set of the ImageNet for evaluation. 
We want to evaluate which network contains the richest information regarding linguistic embedding.
Interestingly, we can go beyond such an evaluation, instead, asking {\it the network $\Theta_X$ to learn concepts 
above and beyond what the network $\Theta_Y$ has learned}. To do so, we include the partial distance correlation into the loss. Unlike the experiment discussed above (minimizing distance correlation), in this setup, we  seek to maximize partial distance correlation. 
The $\text{Loss}_{\text{total}}$ is 
\vspace{-0.5em}
\begin{align}
     \text{Loss}_{\text{CE}}(f_1(x),y) - \alpha \cdot \text{Loss}_{\text{PDC}}\left( (g_1(x)| g_2(x)), gt \right)
    \label{pdc_loss}
\end{align}
\vspace{-2em}

We take pretrained networks $\Theta_X,\Theta_Y$ and then finetune $\Theta_X$ using \eqref{pdc_loss}. The learning rate is set to be $1e-5$ and $\alpha$ in the loss term is $1$. 
To check the benefits of partial DC, we use Grad-CAM \cite{selvaraju2017grad} to highlight the areas that each network is looking at, together with what $\Theta_X$ conditioned on $\Theta_Y$ sees then.

\noindent{\bf Results (a).}
We first show information comparison between two networks. The details of DC and  partial DC are shown in Table. \ref{tab:pdc}. The reader will notice that since ViT achieves the best test accuracy, it also contains the most information. Additionally, although better test accuracy normally coincides with more information, this is not always true. Resnet 50 contains more linguistic information than the much deeper Resnet 152, perhaps a compensation mechanism. For Resnet 152, the network is deep enough to focus on local structures that overwhelm the linguistic information (or this information is unnecessary). 
This experiment suggests a new strategy to compare two networks beyond test accuracy. 

\begin{figure}[!b]
\vspace{-1em}
        \centering
            \includegraphics[width=0.92\columnwidth]{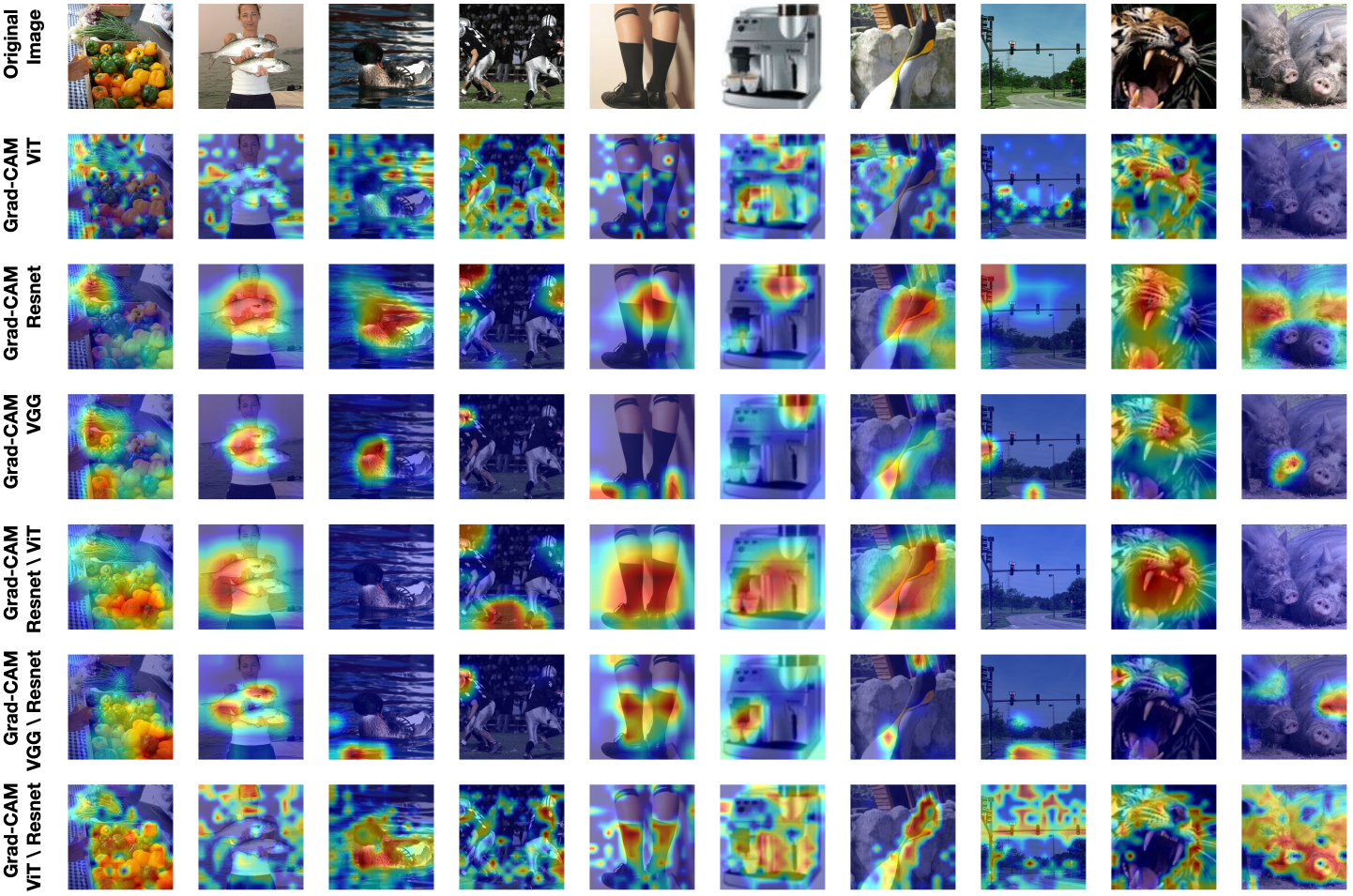}
\vspace{-1em}
               \caption{\footnotesize Grad-CAM results on ImageNet using ViT, Resnet18 and VGG16. After using Partial DC to remove the information learned by another network, ViT can focus on detail places and Resnet can only look in major spots. Similar issue happens to VGG. }
               \label{fig:gradcam}
% \vspace{-1em}
\end{figure}

\noindent{\bf Results (b).} After using a pretrained network, we can also check that by including the partial distance correlation in the loss, which regions does the model pay attention to, using Grad-CAM. 
% The only difference from Grad-CAM is that instead of using the intensity of a single class in the last layer, we use the partial distance correlation as the loss term. 
We replace the loss term of Grad-CAM with the partial distance correlation.
% Everything else remains unchanged. 
The results are shown in Fig. \ref{fig:gradcam}. We see that the pretrained ViT sees across the whole image in different locations, while the Resnet (VGG) tends to focus on only one area of the image. After training, ViT (conditioned on Resnet) pays more attention to the subjects, especially locations outside the Resnet focus. Such experiments help understand how ViT learns {\em beyond} Resnets (CNNs).

\section{Disentanglement}
% {\color{red}
% Follow the paper: An Image is Worth More Than a Thousand Words: 
% Towards Disentanglement in the Wild in NeurIPS 2021. 
% }
\noindent{\bf Overview.}
This experiment studies disentanglement \cite{higgins2016beta,kim2018disentangling,chen2018isolating,locatello2019disentangling,gabbay2021image}. It is believed that the image data are generated from low dimensional latent variables -- but isolating and disentangling the latent variables is challenging. A key in disentangled latent variable learning is to make the factors in the latent variables independent \cite{akash2021learning}. Distance correlation fits perfectly and can handle a variety of dimensions for the latent variables. When the distance correlation is $0$, we know that the two variables are independent. 

\noindent{\bf Experimental settings.}
We follow \cite{gabbay2021image} which focuses on semi-supervised disentanglement to generate high-resolution images. In \cite{gabbay2021image}, one divides the latent variables into two categories: (a) attributes of interest -- a set of semantic and interpretable attributes, e.g. hair color and age; (b) residual attributes -- the remaining information. Formally, $x_i=G(f_i^1,...,f_i^k,r_i)$, where $G$ is the generator that uses the factors of interest $f_i^l$ and the residual to generate image $x_i$. 

In order to enforce the condition that the information regarding the attributes of interest is not leaking into the residual representations, the authors of \cite{gabbay2021image} introduced the loss $\text{L}_{\text{res}}=\sum_{i=1}^n ||r_i||^2$ to limit the residual information. This is sub-optimal as there can be cases where $r_i$ is not $0$ but still independent to the factors of interest $(f_i^l)_{l=1}^k$. Thus, we use  distance correlation to replace this loss:
\vspace{-2em}
\begin{align}
    \text{L}_{\text{res}}=dCor([f^1;f^2;...;f^k], r)
    \label{eq:disent}
\end{align}
\vspace{-1.5em}

We use the same structure proposed in \cite{gabbay2021image}, while the generator architecture is adopted from StyleGAN2 \cite{karras2020analyzing}. The dataset is the human face dataset FFHQ \cite{karras2019style}, and the attributes are: age, gender, etc. We use CLIP \cite{radford2021learning} to partially label the attributes to generate the semi-supervised dataset for training. 
All losses from \cite{gabbay2021image} are used, except that $\text{L}_{\text{res}}$ is replaced by \eqref{eq:disent}.
% The detailed information can be found in the supplementary document.

\begin{figure}[!b]
\vspace{-1em}
        \centering
               \includegraphics[width=0.84\columnwidth]{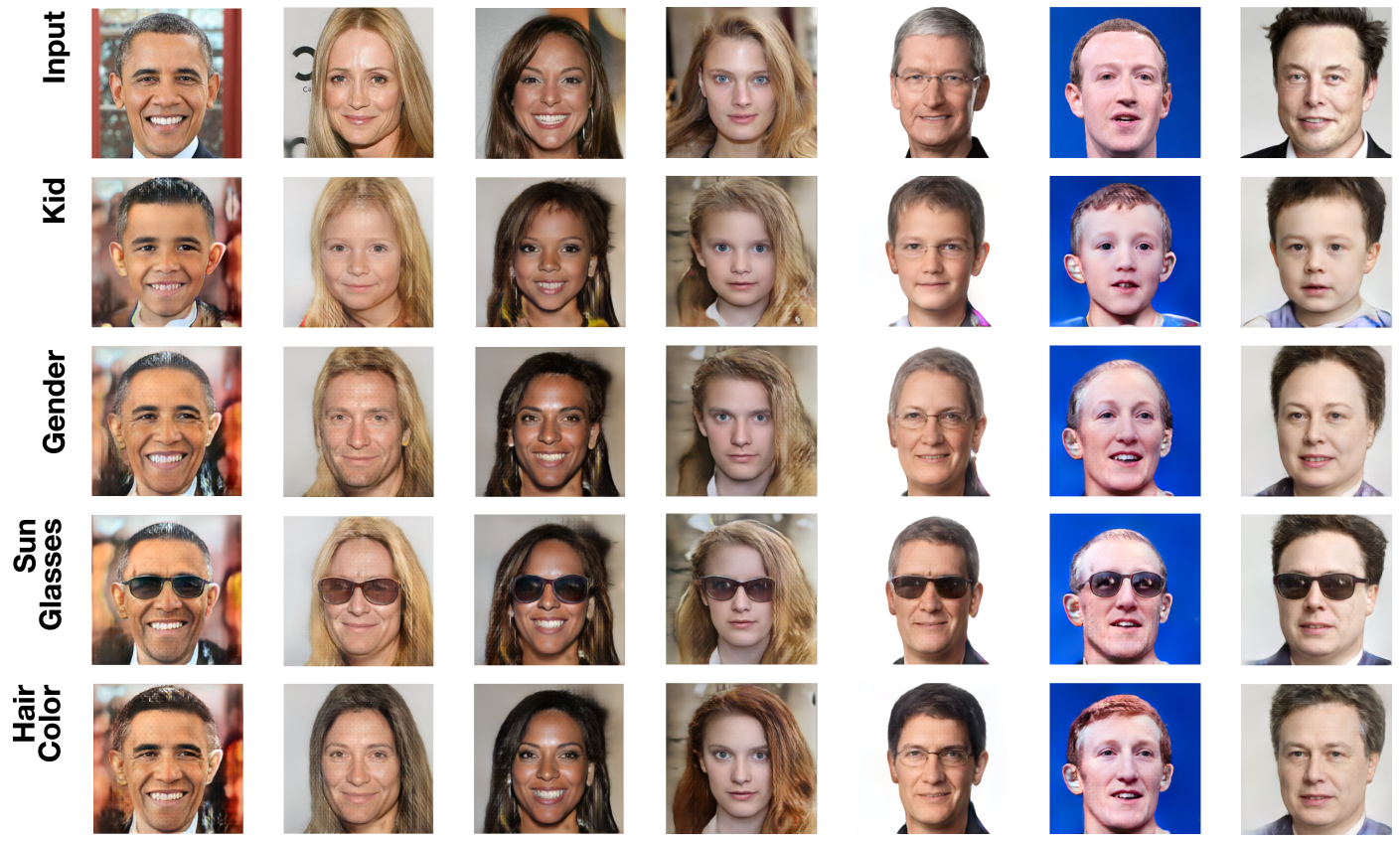}
\vspace{-1em}
               \caption{\footnotesize Representative generated images using our training on FFHQ. Note that these results only use semi-supervised dataset by CLIP. Our methods shows the ability to disentangle the attributes of interest and the remaining information. }
               \label{fig:disent}
%\vspace{-1em}
\end{figure}

\noindent{\bf Results.} 
(Shown in Fig. \ref{fig:disent}) Our model shows the ability to change specific attributes without affecting residual features, such as posture (also see supplement). 
If you use distance correlation for disentanglement, please give credit to the following paper \cite{liu2020metrics} which discusses a nice demonstration of distance correlation helps content style disentanglement. We were not aware of this paper when we wrote the paper last year and thank Sotirios Tsaftaris for communicating his findings with us.

\section{Conclusions}

In this paper, we studied how distance correlation (and partial distance correlation) has a wide variety of uses in deep learning tasks in vision. The measure offers various properties 
that are often enforced using alternative means, that are often 
far more involved. Further, it is extremely simple to incorporate in contrast to various divergence-based measures often used in invariant representation learning.
Notably, the use of partial distance correlation offers the ability of conditioning, which is underexplored in the community. 
We showcase three very different settings, ranging from network comparison to training distinct/different models to disentanglement where the idea is immediately beneficial, and expect that numerous other applications will emerge in short order.   

\smallskip
\noindent {\bf Acknowledgements.} Research supported in part by NIH grants RF1 AG059312,  RF1 AG062336, and RF1AG059869, and NSF 
grant CCF \#1918211. \\
The authors thank Grace Wahba for introducing us to the Gabor Szekely's work on distance correlation in 2014. 

\clearpage
\bibliographystyle{splncs04}
\bibliography{egbib}

\renewcommand\thesection{\Alph{section}}
\renewcommand*\theequation{{\Alph{section}(\arabic{equation})}}
\renewcommand\thefigure{\thesection.\arabic{figure}}  
\renewcommand\thetable{\thesection.\arabic{table}} 
\setcounter{section}{0}
\setcounter{algocf}{0}
\setcounter{equation}{0}
\setcounter{proposition}{0}
\setcounter{figure}{0}
\setcounter{table}{0}

\clearpage
\section{Technical Analysis Using Block Stochastic Gradient}
In this section, we describe results (which were briefly mentioned in the main paper)
showing the viability of a stochastic 
scheme for using distance correlation within 
the loss when training our neural network models. 

In the paper, we noted the existence 
of an algorithm where the convergence rate of the stochastic version of distance correlation in the deep neural network setting is $O(\frac{1}{\sqrt{T}})$. 
Here, we will describe it more formally. 

We note that 
SGD works well for the distance correlation objective -- and so a majority 
of users will revert to such mature implementations 
anyway. Despite desirable practical behavior, its theoretical analysis of the form 
included here is more involved. Therefore, 
the analysis shown below, a modified version 
of \cite{xu2015block}, is reassuring in the sense that we know that stochastic 
updates (carried out in a specific way) can provably optimize our loss. 

\subsection{Notation in the Proof}
We use $\Theta_X,\Theta_Y$ to denote the parameters of neural networks, and $X,Y$ as features extracted by the respective neural networks. Let the minibatch size be $m$, and the dataset $\mathcal{D}=(\mathcal{D_X},\mathcal{D_Y})$ be of size $n$. 
Let, $X\in \mathbb{R}^{m\times p}, Y \in \mathbb{R}^{m\times q}$, with $p,q$ be the dimension of features. 
We use $(x_t,y_t)_{t=1}^T,x_t\subset \mathcal{D_X},y_t\subset \mathcal{D_Y}$ to represent the data samples at step $t$, $T$ is the total number of training steps. The distance matrices $A_t,B_t$ are computed when given $X_t,Y_t$ using \eqref{eq:DC_A}, which is of dimension $m\times m$ for each minibatch. Further, we use $(X_t)_k$ to represent the $k^{th}$ element in $X_t$. Also,  $(A_t)_{k,l}$ is the  $k^{th}$ row and $l^{th}$ column element in the matrix $A_t$.
The inner-product between two matrices $A,B$ is defined as $\langle A,B \rangle = \sum_{i,j}^{m} (A)_{i,j}(B)_{i,j}$.

\begin{align}
    &a_{k,l} = \|X_k-X_l\|, \quad \bar{a}_{k, \cdot}=\frac{1}{n}\sum_{l=1}^n a_{k,l}, \quad \bar{a}_{\cdot, l}=\frac{1}{n}a_{k,l}, \nonumber \\[-1em]
    &\bar{a}_{\cdot, \cdot} = \frac{1}{n^2} \sum_{k,l=1}^n a_{k,l}, \quad A_{k,l} = a_{k,l} - \bar{a}_{k,\cdot} -\bar{a}_{\cdot, l} + \bar{a}_{\cdot, \cdot} \label{eq:DC_A}
\end{align}

\subsection{Objective Function}
Consider the case where we minimize DC between 
two networks $\Theta_X,\Theta_Y$. Since the parameters between $\Theta_X,\Theta_Y$ are separable, we can use block stochastic gradient iteration in \cite{xu2015block} with some modification.

To minimize the distance correlation, we need to solve the following problem
\begin{align}
    \min_{\Theta_X,\Theta_Y} & \ & \frac{\langle A(\Theta_X;x),B(\Theta_Y;y) \rangle}{\sqrt{\langle A(\Theta_X;x),A(\Theta_X;x)\rangle \langle B(\Theta_Y;y),B(\Theta_Y;y)\rangle}} \\
    \mbox{s.t.} & \ & (A)_{k,l} =  || (X)_k - (X)_l ||_2,\: X = \Theta_X(x)\\ 
    \ & \ & (B)_{k,l} = || (Y)_k - (Y)_l ||_2, \:\: Y = \Theta_Y(y) \nonumber
\end{align}
We slightly abuse the notation of $\Theta_X(x)$ to correspond to applying the network $\Theta_X$ on the data $x$, and reuse $A$ to simplify the notation $A(\Theta_X;x)$ and the distance matrix.
We can rewrite the expression (with $A$, $B$ defined above) using:
\begin{align}
    \min_{\Theta_X,\Theta_Y} \langle A, B \rangle \:\: \text{s.t.} \:\: & \max_{x\subset \mathcal{D_X}}\langle A,A\rangle\leq m; \:\: \max_{y\subset \mathcal{D_Y}}\langle B,B\rangle \leq m 
    % & \max_{y_t}\|B\|_{*}\leq m; \:\: \max_{y_t}\|B\|_2 \leq \beta/\sqrt{m}
\end{align}
where $(x,y)$ are the minibatch of samples from the data space $(\mathcal{D_X},\mathcal{D_Y})$.

% In the block stochastic gradient setup, we rewrite the following equation,
We can  rewrite as the following expression similar to Eq. (1) in \cite{xu2015block}.
\begin{align}
    \min_{\Theta_X,\Theta_Y}\Phi(\Theta_X,\Theta_Y) = \mathbb{E}_{x,y}f(\Theta_X,\Theta_Y;x,y) + \gamma(\Theta_X) + \gamma(\Theta_Y)
\label{eq:minPhi}
\end{align}
where $f(\Theta_X,\Theta_Y;x,y)$ is $\langle A, B \rangle$ and $\gamma(\Theta_X)$ encodes the convex constraint on the network $\Theta_X$, i.e.,  $\max_{x\subset\mathcal{D_X}}\langle A,A\rangle\leq m$. Similarly, $\gamma(\Theta_Y)$ encodes $\max_{y\subset \mathcal{D_Y}}\langle B,B\rangle \leq m$. Further, $\Phi(\Theta_X,\Theta_Y)$ is the constrained objective function to be optimized. 

\subsection{Block Stochastic Gradient Iteration}
We adapt Alg. 1 from \cite{xu2015block} to our case in Alg. \ref{alg:bsg}. Since we will need the entire minibatch $(x_t,y_t)$ to compute the objective function, there will be no mean term when computing the sample gradient $\tilde {\mathbf{g}}_{X}^{t}$. Further, since both blocks $(\Theta_X,\Theta_Y)$ are constrained, line $3,5$ will use (5) from \cite{xu2015block}. The detailed algorithm is presented in Alg. \ref{alg:bsg}. 

\renewcommand{\algorithmicrequire}{\textbf{Input:}}
\renewcommand{\algorithmicensure}{\textbf{Output:}}
\begin{algorithm*}
\caption{Block Stochastic Gradient for Updating Distance Correlation}\label{alg:bsg}
    \begin{algorithmic}[1]
        \REQUIRE Two neural network with starting point $\Theta_X^1,\Theta_Y^1$. Training data $\{(x_t,y_t)\}_{t=1}^T$, step size $\eta_X,\eta_Y$, and batch size $m$.
        \ENSURE $\tilde \Theta_X^T,\tilde \Theta_Y^T$
        \FOR {$t=1,\cdots,T$}
            \STATE Compute sample gradient for $\Theta_X$ \\
             \:\: $\tilde{\mathbf{g}}_{X}^{t} = \nabla_{\Theta_X} f(\Theta_X^t,\Theta_Y^t;x_t,y_t) $
            \STATE $\Theta_X^{t+1}=\argmin_{\Theta_X} \langle \tilde{\mathbf{g}}_{X}^{t} + \tilde{\nabla} \gamma_{X}(\Theta_X^t), \Theta_X-\Theta_X^t \rangle + \frac{1}{2\eta_X}\| \Theta_X - \Theta_X^t \|^2 $
            \STATE Compute sample gradient for $\Theta_Y$ \\
             \:\: $\tilde{\mathbf{g}}_{Y}^{t} = \nabla_{\Theta_Y} f(\Theta_X^{t+1},\Theta_Y^t;x_t,y_t) $
            \STATE $\Theta_Y^{t+1}=\argmin_{\Theta_Y} \langle \tilde{\mathbf{g}}_{Y}^{t} + \tilde{\nabla} \gamma_{Y}(\Theta_Y^t), \Theta_Y-\Theta_Y^t \rangle + \frac{1}{2\eta_Y}\| \Theta_Y - \Theta_Y^t \|^2 $
            % \STATE $X_{t+1} \leftarrow X_t + \alpha \eta {d_X}_t$ \COMMENT{\% $\alpha$ is chosen as $\min_{t,k,l} ({\|\frac{dA}{d{X_t}_k} - \frac{dA}{d{X_t}_l}\|})^{-1}$ {\color{red}{We need $\|A(X_t+\alpha \eta \partial_t \frac{dA}{dX})\| \leq \|A(X_t)+\eta \partial_t\| $}}}
        \ENDFOR
    \STATE $\tilde \Theta_X^T= \frac{1}{T} \sum_{t=1}^{T}\Theta_X^t$
    \STATE $\tilde \Theta_Y^T= \frac{1}{T} \sum_{t=1}^{T}\Theta_Y^t$
    \end{algorithmic}
\end{algorithm*}

\begin{proposition}
    After $T$ iterations of Algorithm \ref{alg:bsg} with step size $\eta_X=\eta_Y =\frac{\eta}{\sqrt{T}} < \frac{1}{L}$, for some positive constant $\eta<\frac{1}{L}$, where $L$ is the Lipschitz constant of the partial gradient of $f$, by Theorem. 6 in \cite{xu2015block}, we know there exists an index subsequence $\mathcal{T}$ such that:
    \begin{align}
        \lim_{t\rightarrow \infty,t\in \mathcal{T}} \mathbb{E}[\text{dist}(\mathbf{0}, \nabla\Phi(\Theta_X^t,\Theta_Y^t))]=0
    \end{align}
    where $\text{dist}(\mathbf{y},\mathcal{X})=\min_{\mathbf{x}\in \mathcal{X}}\| \mathbf{x}-\mathbf{y}\|$. 
    
    Further, in the special case where  $\mathbb{E}_{x,y}f(\Theta_X,\Theta_Y;x,y)$ is convex, by Theorem. 1 in \cite{xu2015block}, the following statement holds:
    {\small
    \begin{align}
        \mathbb{E}[\Phi(\tilde\Theta_X^T,\tilde\Theta_Y^T) - \Phi(\Theta_X^*,\Theta_Y^*)] &\leq D\eta \frac{1+\log T}{\sqrt{1+T}} + \frac{\| \Theta_X -\Theta_X^1 \|^2 + \| \Theta_Y -\Theta_Y^1 \|^2}{2\eta \sqrt{1+T}}
    \end{align}
    }%
    where $\tilde \Theta_X,\tilde \Theta_Y$ is computed in Algorithm \ref{alg:bsg}, $\Theta_X^*,\Theta_Y^*$ are the optimum of the desired function, and $D$ is a constant depending on $\|( \Theta_X^*;\Theta_Y^*) \|$. 
\label{prop:bound}
\end{proposition}    

\subsection{Modification from BSG \cite{xu2015block} to DC}
The statement of Prop. \ref{prop:bound} is similar to the statement of Theorem 1 and 6 in \cite{xu2015block}. So, we can use the statement in \cite{xu2015block} with some modification of our setup. 
We define 
\begin{align*}
    F(\Theta_X,\Theta_Y) = \mathbb{E}_{x,y}f(\Theta_X,\Theta_Y;x,y),\:\: \Gamma(\Theta_X,\Theta_Y)=\gamma(\Theta_X)+\gamma(\Theta_Y)
\end{align*}

Then, for the gradient w.r.t. $X$, we have the following expression (similar for $Y$):
\begin{align*}
    \tilde{\mathbf{g}}_{X}^{t} &= \nabla_{\Theta_X} f(\Theta_X^{t},\Theta_Y^t;x_t,y_t)\\
    {\mathbf{g}}_{X}^{t} &= \nabla_{\Theta_X} F(\Theta_X^{t},\Theta_Y^t)\\
    \boldsymbol{\delta}_X^t &= \tilde{\mathbf{g}}_{X}^{t} - {\mathbf{g}}_{X}^{t}
\end{align*}

We first restate four assumptions from \cite{xu2015block}.
\begin{assumption}
    There exist a constant $c$ and a sequence $\{\sigma_k\}$ such that for any $t$, 
    \begin{align*}
        & \| \mathbb{E}[\boldsymbol{\delta}_X^t| x_t,y_t ] \| \leq c\cdot \max(\eta_X,\eta_Y),\\
        & \mathbb{E}\| \boldsymbol{\delta}_X^t \|^2 \leq \sigma_t^2
    \end{align*}
    \label{as:1}
\end{assumption}

\begin{assumption}
    The objective function is lower bounded, i.e., $\Phi(\Theta_X,\Theta_Y) > -\infty$. And there is a uniform Lipschitz constant $L>0$ such that:
    \begin{align*}
        \| \nabla_X F(\Theta_X,\Theta_Y) - \nabla_X F(\Theta_X',\Theta_Y') \| \leq L \| (\Theta_X;\Theta_Y) - (\Theta_X';\Theta_Y') \|, \\
        \forall (\Theta_X,\Theta_Y),(\Theta_X',\Theta_Y')
    \end{align*}
    \label{as:2}
\end{assumption}

\begin{assumption}
    There exists a constant $\rho$ such that $\mathbb{E} \| (\Theta_X^t;\Theta_Y^t) \|^2 \leq \rho^2$ for all $t$.
\label{as:3}
\end{assumption}

\begin{assumption}
    The constraint function $\gamma$ is Lipschitz continuous. There is a constant $L_\gamma$, such that:
    \begin{align*}
        \| \gamma(\Theta_X) - \gamma(\Theta_X') \| \leq L_{\gamma} \| \Theta_X-\Theta_X' \|, \forall \Theta_X,\Theta_X'
    \end{align*}
    \label{as:4}
\end{assumption}

\setcounter{theorem}{5}
\begin{theorem}
    (from \cite{xu2015block}) Let $\{\Theta^t\}$ be generated from Algorithm \ref{alg:bsg} with $\eta^t_X, \eta^t_Y$, being constained as,
    \begin{align*}
        0 < \inf_t \eta_X^t \leq \sup_t \eta_X^t < \frac{1}{L}\\
        0 < \inf_t \eta_Y^t \leq \sup_t \eta_Y^t < \frac{1}{L}\\
    \end{align*}
    Under Assumptions \ref{as:1} through \ref{as:4}, if either $\mathcal{X}=\mathbb{R}^{n_X},\mathcal{Y}=\mathbb{R}^{n_Y}$ or $\gamma=0$, and 
    \begin{align*}
        \sum_{t=1}^\infty \sigma_t^2 < \infty
    \end{align*}
    then there exists an index subsequence $\mathcal{T}$ such that
    \begin{align*}
        \lim_{t\rightarrow \infty,t\in \mathcal{T}} \mathbb{E}[\text{dist}(\mathbf{0}, \nabla\Phi(\Theta_X^t,\Theta_Y^t))]=0, 
    \end{align*}
    where $\text{dist}(\mathbf{y},\mathcal{X})=\min_{\mathbf{x}\in \mathcal{X}}\| \mathbf{x}-\mathbf{y}\|$.
\end{theorem}

\begin{remark}
    In our case, we have that 
    $\Theta_X,\Theta_Y$ are the parameters of neural networks. During training, we have no constraint on the weights and biases, so the space of $\Theta_X$, which is $\mathcal{X}$, is the Euclidean space. 
    Also, we can have $\eta_X^t=\eta_Y^t=\frac{\eta}{\sqrt{t}}<\frac{1}{L}$. 
    All the other assumptions are similar to \cite{xu2015block}. Thus, we have the same result in Prop. \ref{prop:bound}.
\end{remark}

In the convex case, we use the Theorem 1 from \cite{xu2015block}.
\setcounter{theorem}{0}
\begin{theorem}
    \cite{xu2015block} (Ergodic convergence for non-smooth convex case). Let $\{\Theta^t\}$ be generated from Algorithm \ref{alg:bsg} with $\eta^t_X=\eta^t_Y=\eta_t=\frac{\eta}{\sqrt{t}}<\frac{1}{L},\forall t$, for some positive constant $\eta<\frac{1}{L}$. Under Assumptions \ref{as:1} through \ref{as:4}, if $F$ and $\gamma$ are both convex, $\Theta_X^*,\Theta_Y^*$ is a solution of \eqref{eq:minPhi}, and $\sigma=\sup_t \sigma_t < \infty$, then
    \begin{align*}
        \mathbb{E}[\Phi(\Theta_X^t,\Theta_Y^t) - \Phi(\Theta_X^*,\Theta_Y^*)]\leq D\eta \frac{1+\log T}{\sqrt{1+T}} + \frac{\| (\Theta_X^*;\Theta_Y^*) - (\Theta_X^1;\Theta_Y^1) \|^2}{2\eta \sqrt{1+T}}
    \end{align*}
    where $\tilde \Theta_X^T = \frac{\eta_t \Theta_X^{t+1}}{\sum_{t=1}^T \eta_t}$, $\tilde \Theta_Y^T = \frac{\eta_t \Theta_Y^{t+1}}{\sum_{t=1}^T \eta_t}$, and
    \begin{align*}
        D=\frac{s(\sigma^2+4 L_\gamma^2)}{1-L\eta}+\sqrt{s}(\| (\Theta_X^*;\Theta_Y^*) \| + \rho)(c+L\sqrt{8M_\rho^2+8\sigma_t^2+4L_\gamma^2})
    \end{align*}
    where $M_\rho=\sqrt{4L^2\rho^2 + 2\max(\|\nabla_{\Theta_X}F(\mathbf{0})\|^2,\|\nabla_{\Theta_Y}F(\mathbf{0})\|^2)}$
\end{theorem}

\begin{remark}
    Our case is a special case of the Block Stochastic Gradient problem with $s=2$ ($s$ is the number of blocks). So the above theorem can be directly applied to our analysis when $F,\gamma$ are both convex. However, this may not be true in most deep neural networks, we obtain the $O(T^{1/2})$ convergence rate using Algorithm \ref{alg:bsg}. 
\end{remark}

\section{Experimental Details in Section 4: Independent Features Help Robustness}
When we train  $f_1$ using cross entropy loss and $f_2$ using cross entropy loss plus our distance correlation loss (to learn independent features with $f_1$), we first train $f_1$ for one epoch, and then train $f_2$ for one epoch given the current $f_1$, and we repeat this process for the total number of epochs (200 for CIFAR10 and 40 for ImageNet). Our hyperparameter $\alpha$ controls the tradeoff between the cross entropy loss and the distance correlation loss. In practice, we could increase $\alpha$ to emphasize (or weight) learning independent features more, and decrease $\alpha$ if we want to keep the classification accuracy of $f_2$ in standard setting (non adversarial) even closer to that of $f_1$. During training, we do not utilize data augmentation for all experiments. The training is done on Nvidia A100 GPUs. Our distance correlation adds approximtely $20\%$ cost to the training time compared with training only using cross entropy loss.

\begin{figure*}[!tbh]
        \centering
               \includegraphics[width=0.99\columnwidth]{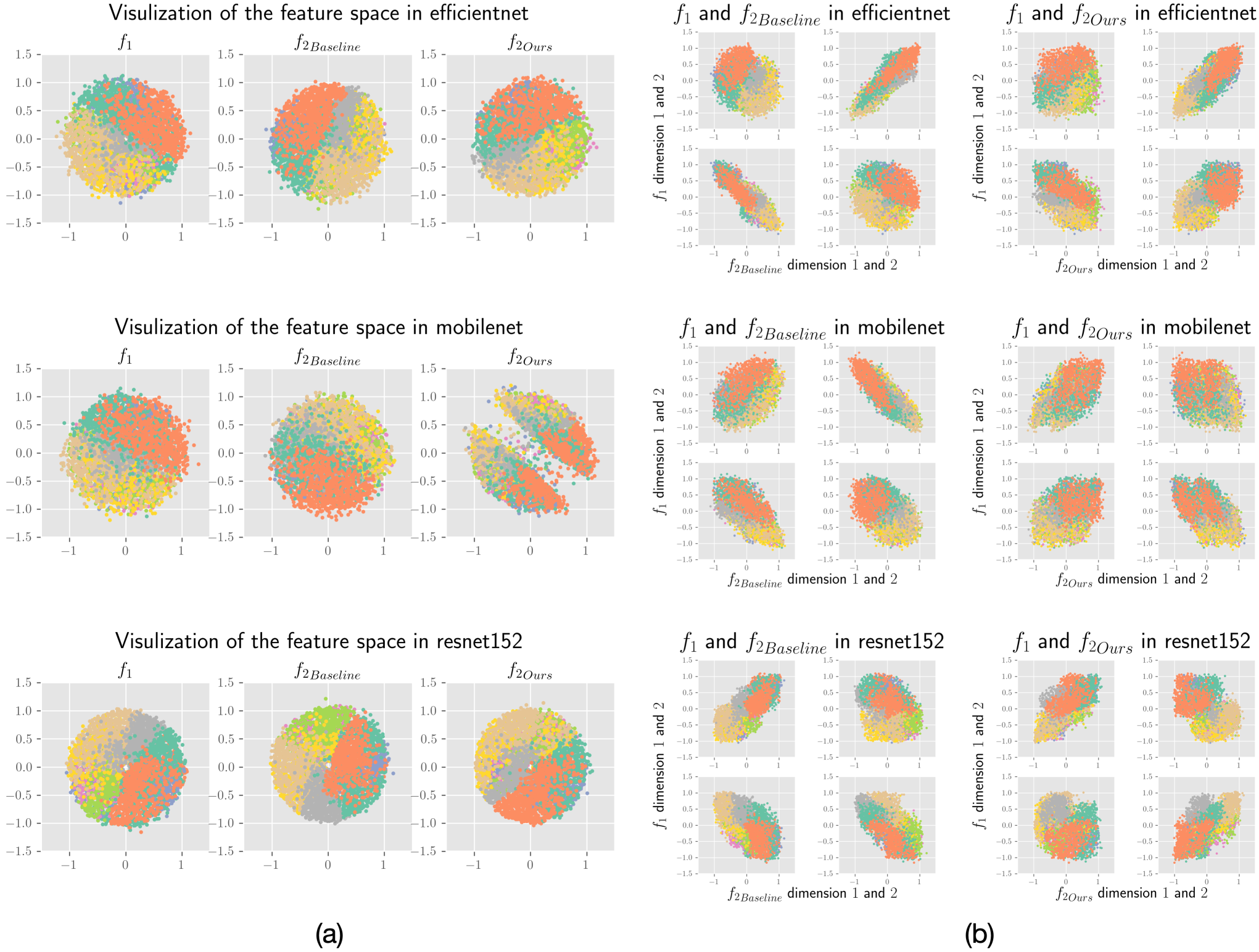}\\
               \caption{\small Picasso visualization of features space and the correlation between different models for all three models. {\bf{(a)}} Feature space distribution. {\bf{(b)}} Cross-correlation between the feature space of $f_1$ and $f_2$ trained with/without DC. We get better independence.}
               \label{fig:diverge}
\end{figure*}

We also include some more visualization of feature spaces in addition to those shown in our main paper in Fig. \ref{fig:diverge}. Our method shows more independence than the baseline model. This implies training with Distance Correlation (DC) can help independence, thus improve robustness to transferred samples.

\section{Experimental Details in Section 5: Informative Comparison Between Networks}

\subsection{Measure similarity between neural networks}
We take the pretrained neural network from \cite{rw2019timm}. The features are reshaped to a 1D vector and we compute the Euclidean distance between samples from the official validation set of ImageNet \cite{krizhevsky2012imagenet}. No finetuning was used in this experiment. The results are shown in the main paper.

\subsection{What remains when ``taking out'' (aka controlling for) $Y$ from $X$}
We will first discuss the details of the heatmap that we plotted using Grad-CAM \cite{selvaraju2017grad}. In the original implementation of Grad-CAM, the model uses one layer as the target and uses both gradients (from the loss function) and the activation in that layer for the visualization. 

In the original Grad-CAM, the loss is extracted as the intensity before the softmax layer of one given class. For example, assume ``dog'' is the $6^{th}$ class in the dataset. If we want to see which location in the image is related to ``dog'', we will use $f(x)[6]$ as the loss function, where $\text{Softmax}(f(x))$ is the final output of the model when given image $x$. 

In our case, the activation remains the same. But the loss function is different. We use the distance correlation between the features extracted by the neural network and the ground truth linguistics embedding, i.e., $\text{Loss}=\mathcal{R}^2(X, GT )$, where $X$ is the feature of input image extracted by the neural network. 

After showing the Grad-CAM results for each individual network, we want to check if the partial distance correlation can help the network focus on a different location. Thus, we finetune the network $X$ with an extra loss term $\text{Loss}_{\text{PDC}}$.

We take ViT-B/16 as our model $X$ and Resnet 18 as our model $Y$. We first load the pretrained weights from \cite{rw2019timm} and finetune model $X$ with model $Y$ being fixed. $\alpha$ in our case is set as $1$ in the loss term 
\begin{align*}
    \text{Loss}_{\text{CE}}(f_1(x),y) - \alpha \cdot \text{Loss}_{\text{PDC}}\left( (g_1(x)| g_2(x)), gt \right)
\end{align*}
Learning rate is set as $1e^{-5}$ and batch size is set as $64$. We train $15$ epochs in total. The finetuning on two RTX 2080Ti takes 2 days. 

\subsection{Extra Results}
\begin{figure}[!thb]
        \centering
            \includegraphics[width=0.92\columnwidth]{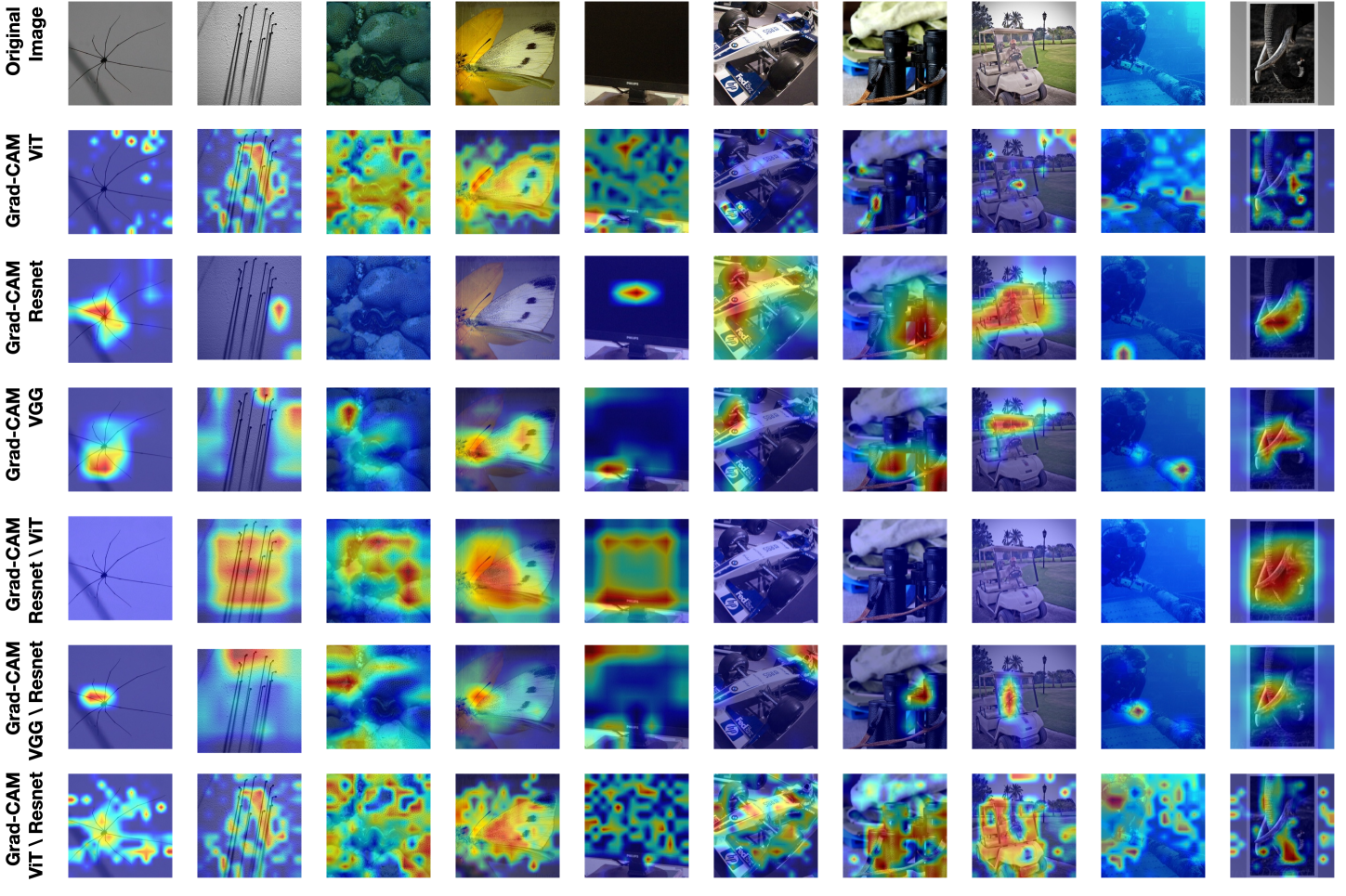}

               \caption{\footnotesize Extra Grad-CAM results on ImageNet using ViT, Resnet18 and VGG16. After using Partial DC to remove the information learned by another network, ViT can focus on detail places and Resnet can only look in major spots. Similar issue happens to VGG. }
               \label{fig:gradcam_extra1}
\end{figure}

\begin{figure}[!hbt]
        \centering
            \includegraphics[width=0.92\columnwidth]{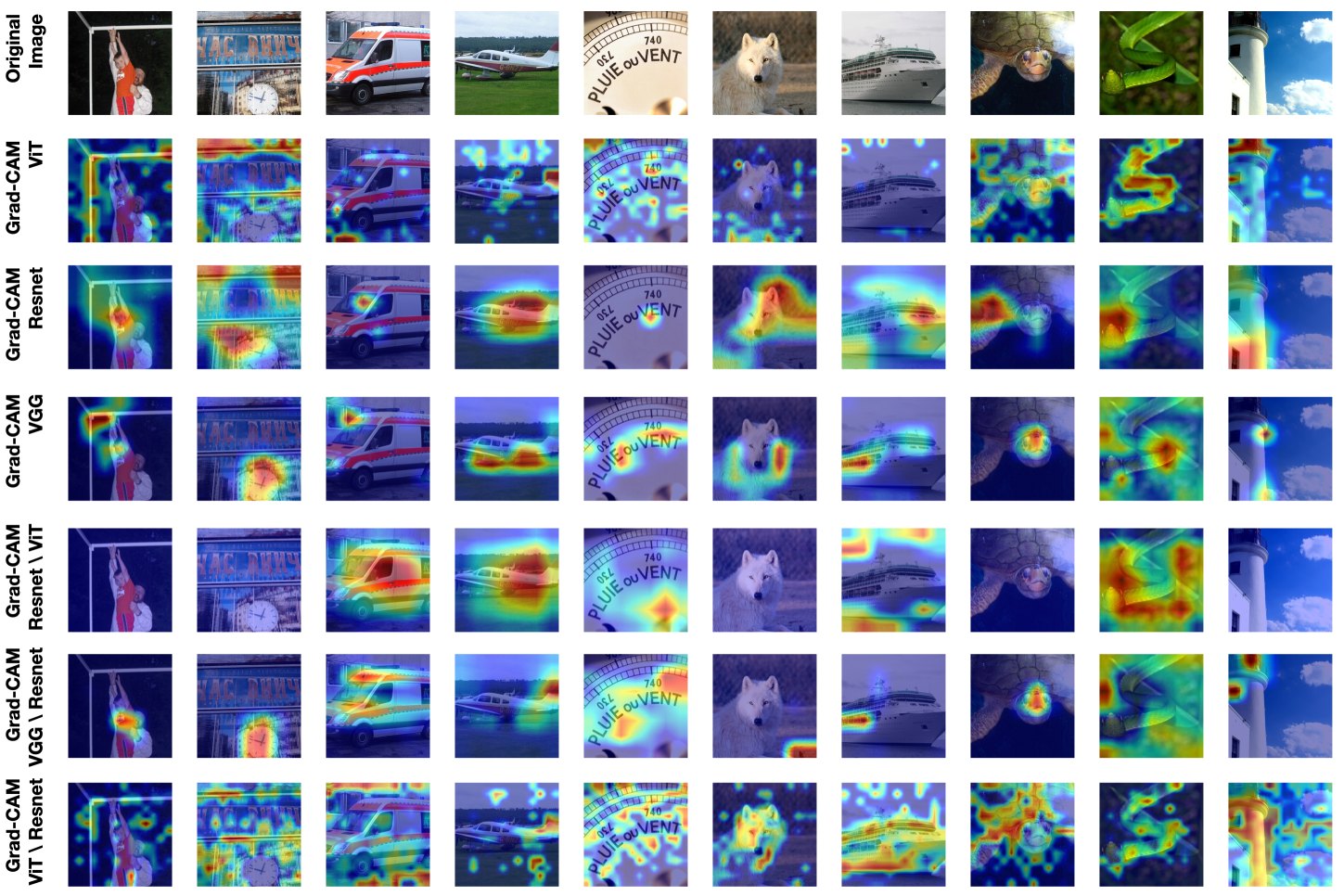}

               \caption{\footnotesize Extra Grad-CAM results on ImageNet using ViT, Resnet18 and VGG16. After using Partial DC to remove the information learned by another network, ViT can focus on detail places and Resnet can only look in major spots. Similar issue happens to VGG. }
               \label{fig:gradcam_extra2}
\end{figure}

We show several additional heat maps using Grad-CAM in addition to those in our main paper in Fig. \ref{fig:gradcam_extra1} and \ref{fig:gradcam_extra2}.

\section{Experimental Details in Section 6: Disentanglement}
We follow the setup in \cite{gabbay2021image} where the dataset contains both labeled and unlabeled data. The provided labels are indicated by the function $\ell$:
\begin{align*}
    \ell(i,j) = \left\{
    \begin{array}{rl}
         1, & f_i^j \text{exists (attribute $j$ of image $i$ is labeled)}\\
        0, & \text{otherwise}
    \end{array}
    \right.
\end{align*}

For each attributes, we train $k$ classifiers of the form $C^j:\mathcal{X}\rightarrow [m^j]$ where $m^j$ denotes the number of values of attribute $j$. The gender attribute here contains male and female, and age attribute contains kid, teenage, adult, and old person. Details are shown in in Table. \ref{tab:attribute}

\begin{table}[!ht]
\caption {\footnotesize Values that we use in the disentangle experiment on FFHQ dataset.} \label{tab:attribute}
\centering
\scalebox{0.85}{
\begin{tabular}{c|c}
\topline\myrowcolour
Attribute & Values\\
age & kid, teenage, adult, old person\\
gender & male, female\\
ethnicity & African person, white person, Asian person\\
hair & brunette, blond, red, white, black, bald\\
beard & beard, mustache, goatee, shaved\\
glasses & glasses, shades, without glasses\\
\bottomline 
\end{tabular}
}
\end{table}

For the classifiers when given the true label, we use the cross-entropy loss\\
\begin{align}
    \text{L}_{cls} &= \sum_{i=1}^n \sum_{j=1}^n \ell(i,j)\cdot H(\text{Softmax}(C^j(x_i) ), f_i^j )
\end{align}

For the classifiers without the true label, we use the entropy so that the information is not leaking
\begin{align}
    \text{L}_{ent} &= \sum_{i=1}^n \sum_{j=1}^n (1-\ell(i,j))\cdot H(\text{Softmax}(C^j(x_i) ) )
\end{align}

For the residual, we use the distance correlation loss in our paper
\begin{align}
    \text{L}_{res}=dCor([f^1;f^2;...;f^k], r)
\end{align}

Let the value for each of the attributes of interest $j$ to be $\tilde f_i^j$
\begin{align*}
    \tilde f_i^j = \left\{
    \begin{array}{ll}
         f_i^j & ,\ell(i,j)=1\\
         \text{Softmax}(C^j(x_i))  & ,\text{otherwise}
    \end{array}
    \right.
\end{align*}

Also, we include the reconstruction loss to generate the target image
\begin{align}
    \text{L}_{rec}=\sum_{i=1}^n \phi (G(\tilde f_i^1, ..., \tilde f_i^k, r_i'), x_i)
\end{align}

The final loss will be the linear combination of all the loss above.
\begin{align}
    \text{L}_{disentangle}=\text{L}_{rec} + \lambda_{cls} \text{L}_{cls} + \lambda_{ent}\text{L}_{ent} + \lambda_{res}\text{L}_{res}
\end{align}

In our implementation, $\lambda_{cls}=0.1$, $\lambda_{ent}=0.01$, $\lambda_{res}=1e^{-5}$. 

\subsection{Additional examples of generated images}
\begin{figure*}[!h]
        \centering
               \includegraphics[height=0.13\columnwidth]{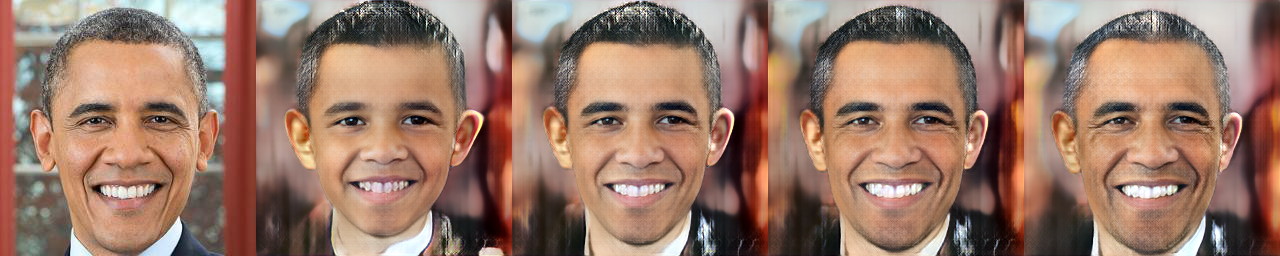}\\
               \includegraphics[height=0.13\columnwidth]{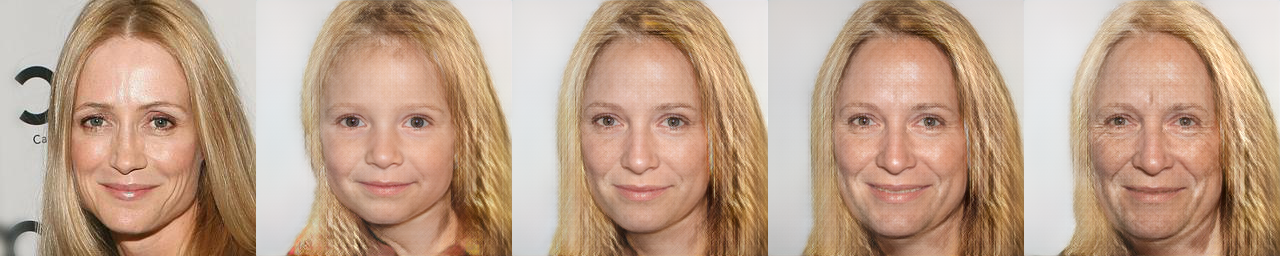}\\
               \includegraphics[height=0.13\columnwidth]{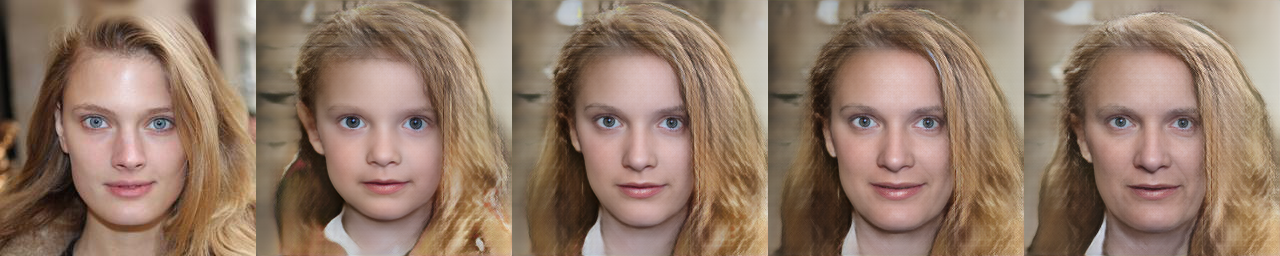}\\
               \includegraphics[height=0.13\columnwidth]{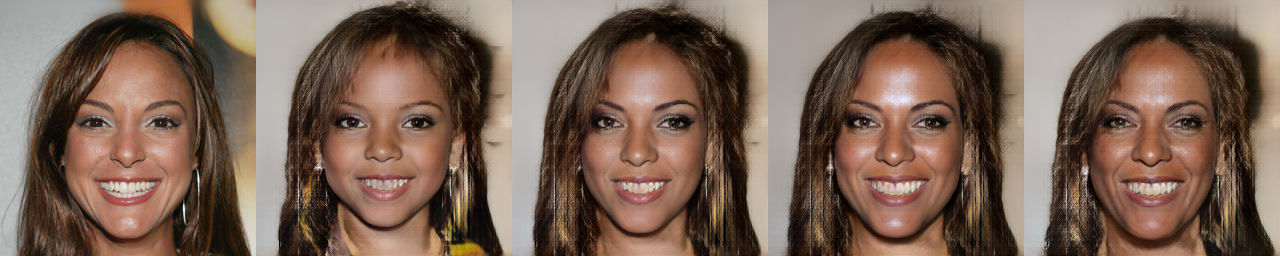}\\
               \includegraphics[height=0.13\columnwidth]{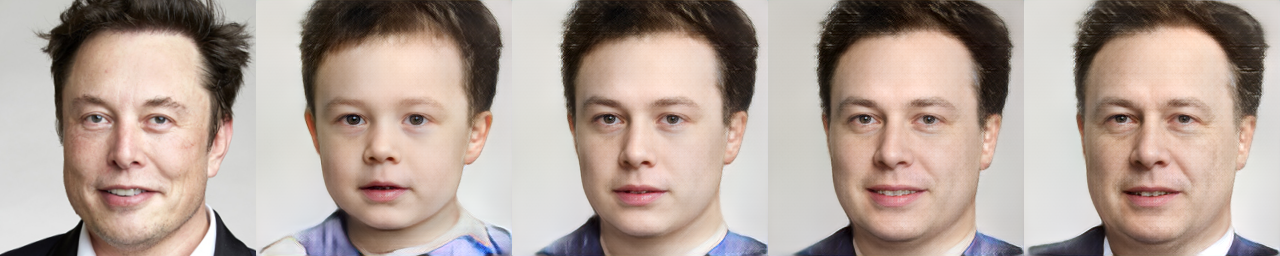}\\
                \includegraphics[height=0.13\columnwidth]{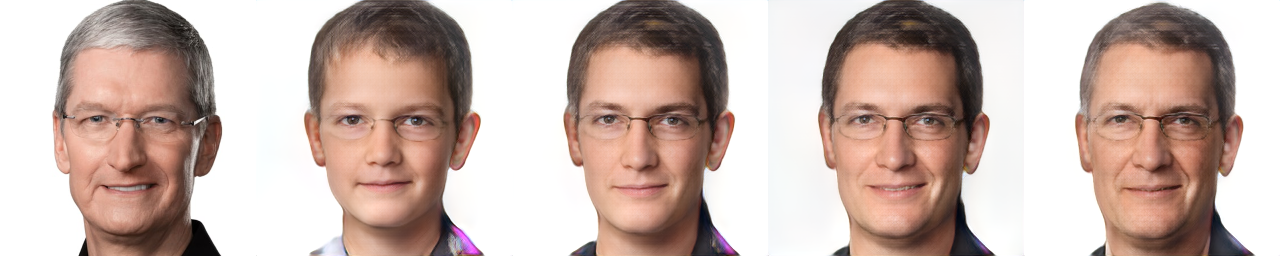}\\
                 \includegraphics[height=0.13\columnwidth]{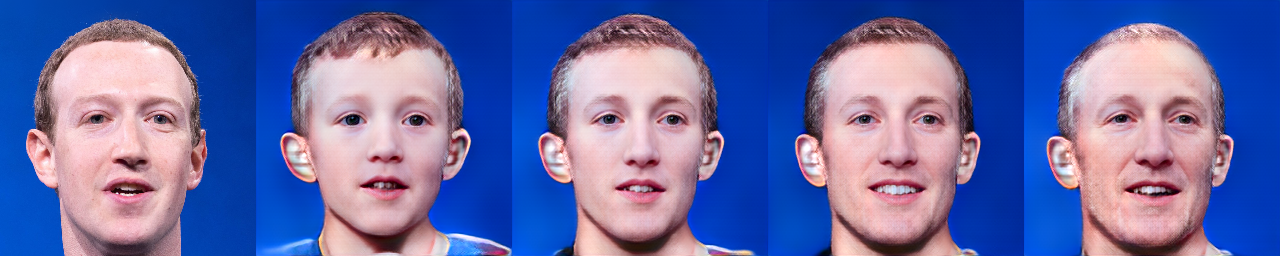}
               \caption{\small  Generated images pertaining to the different ages for the same individual. While the results are qualitative, perceptually the generated results appear meaningful.}
               \label{fig:age}
\end{figure*}

\begin{figure*}[!h]
        \centering
               \includegraphics[height=0.13\columnwidth]{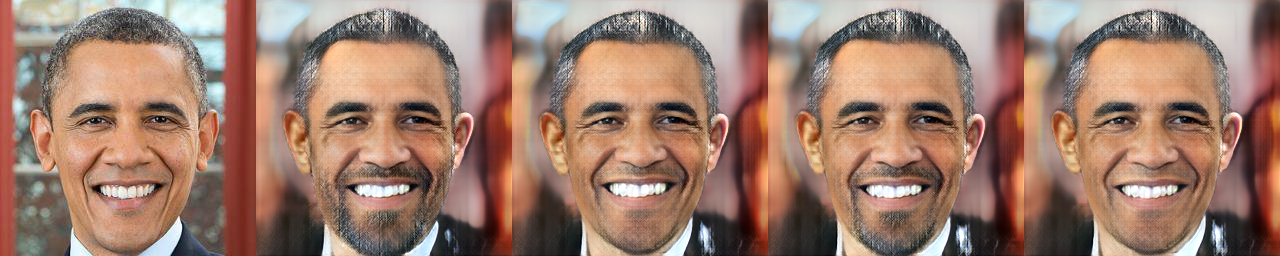}\\
               \includegraphics[height=0.13\columnwidth]{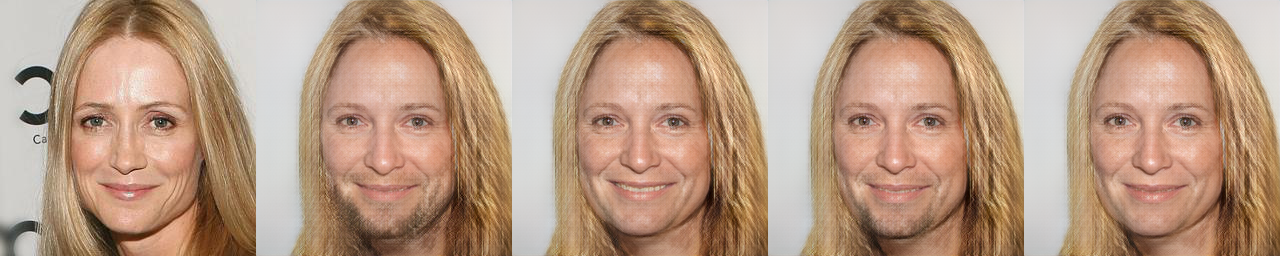}\\
               \includegraphics[height=0.13\columnwidth]{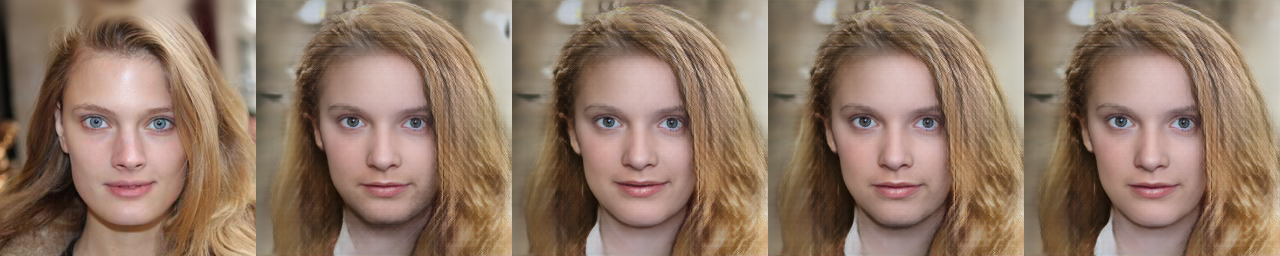}\\
               \includegraphics[height=0.13\columnwidth]{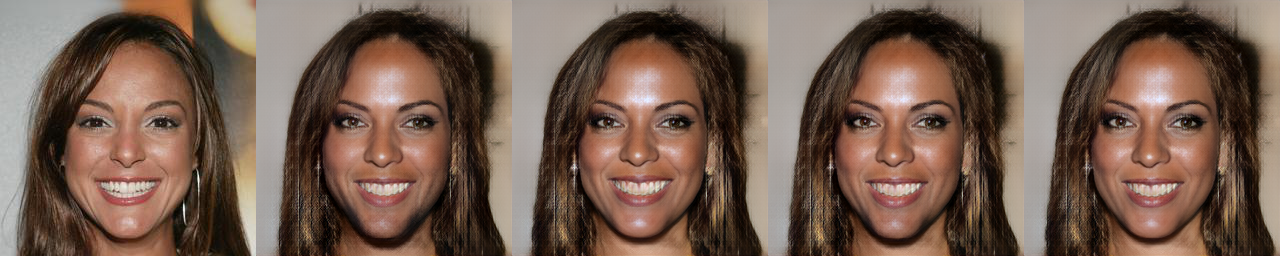}\\
               \includegraphics[height=0.13\columnwidth]{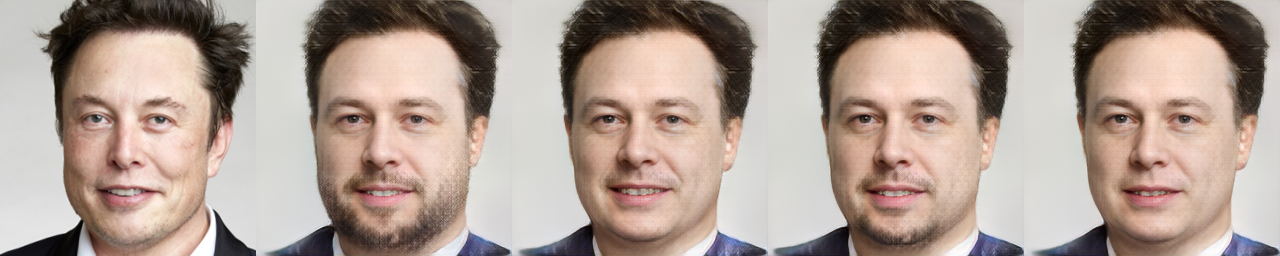}\\
               \includegraphics[height=0.13\columnwidth]{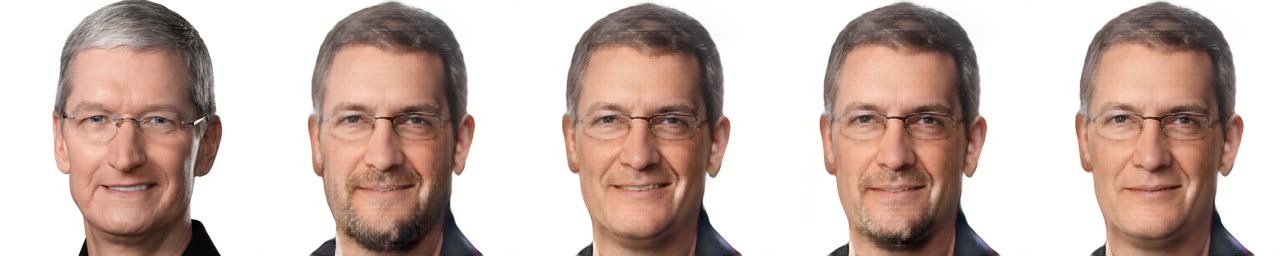}\\
               \includegraphics[height=0.13\columnwidth]{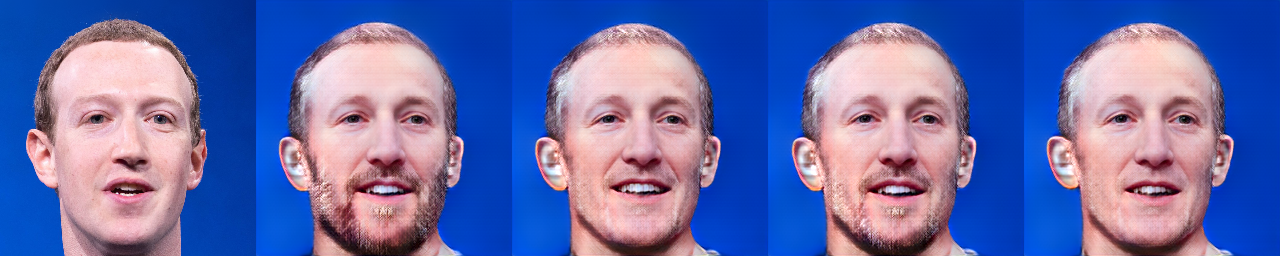}
               \caption{\small Generated images pertaining to different beard levels for the same individual. The first two rows appear perceptually meaningful.}
               \label{fig:beard}
\end{figure*}

\begin{figure*}[!h]
        \centering
               \includegraphics[height=0.13\columnwidth]{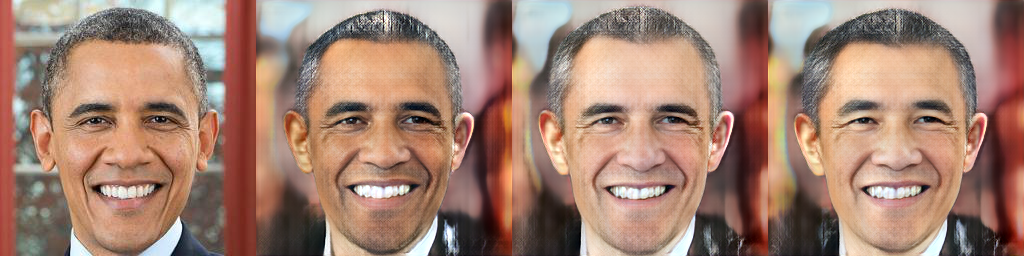}\\
               \includegraphics[height=0.13\columnwidth]{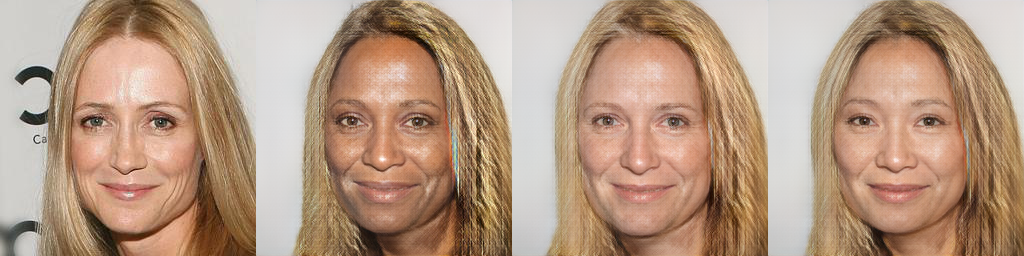}\\
               \includegraphics[height=0.13\columnwidth]{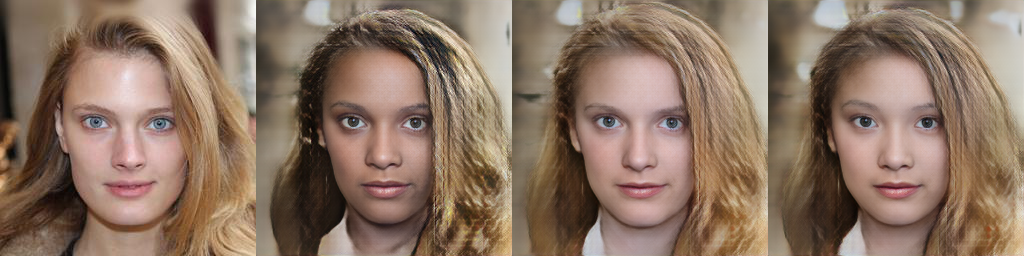}\\
               \includegraphics[height=0.13\columnwidth]{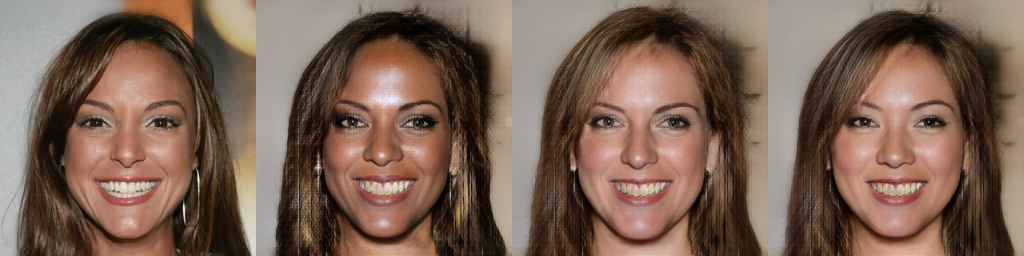}\\
               \includegraphics[height=0.13\columnwidth]{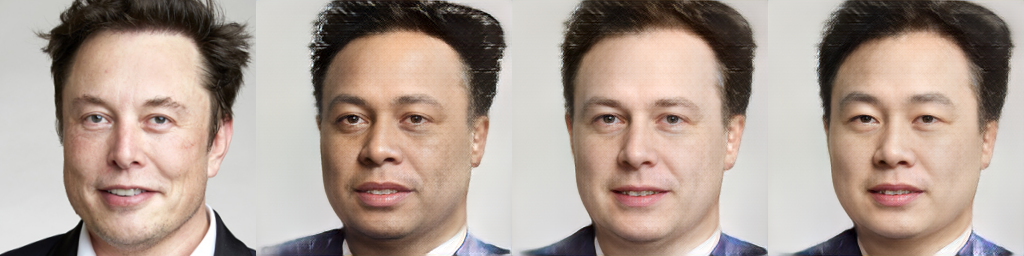}\\
               \includegraphics[height=0.13\columnwidth]{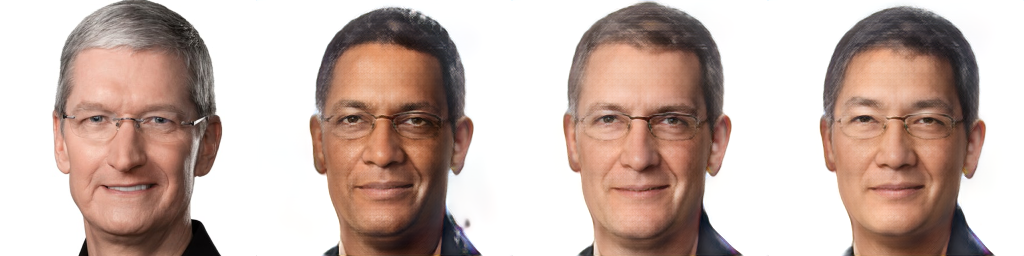}\\
               \includegraphics[height=0.13\columnwidth]{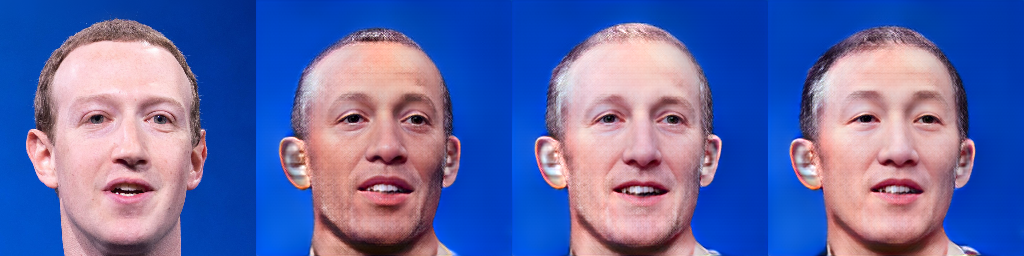}
               \caption{\small  Generated images pertaining to different ethnicity for the same individual.}
               \label{fig:ethnicity}
\end{figure*}

\begin{figure*}[!h]
        \centering
               \includegraphics[height=0.13\columnwidth]{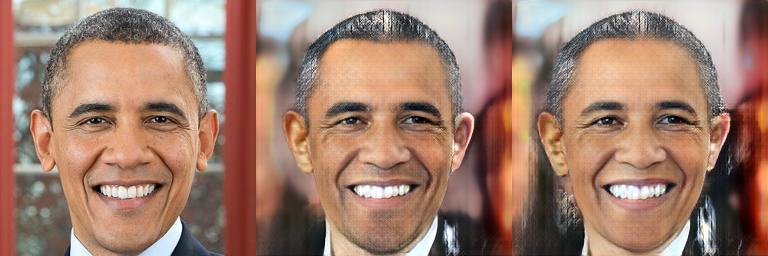}\\
               \includegraphics[height=0.13\columnwidth]{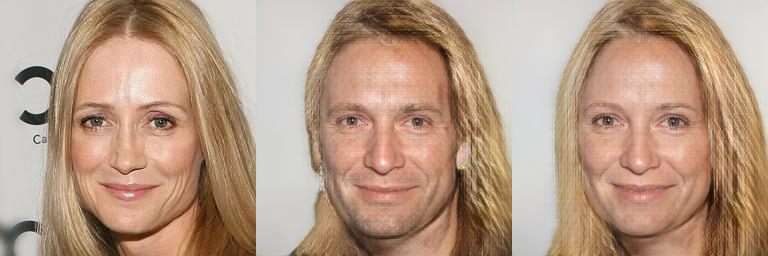}\\
               \includegraphics[height=0.13\columnwidth]{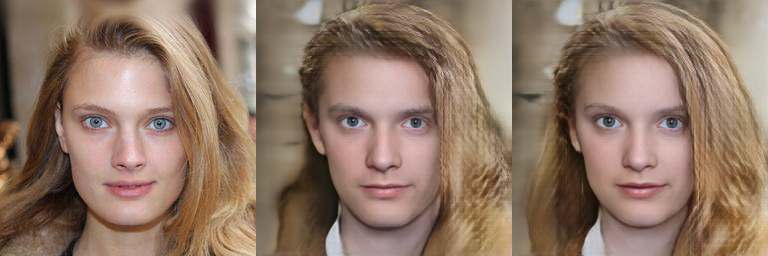}\\
               \includegraphics[height=0.13\columnwidth]{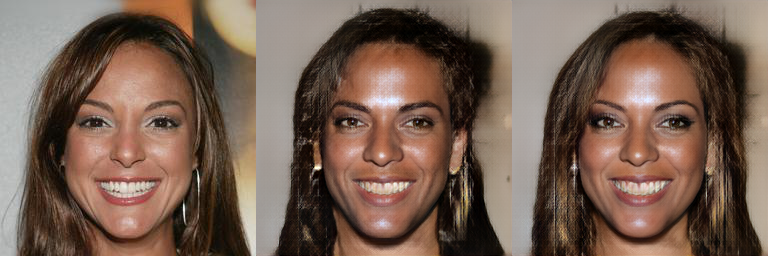}\\
               \includegraphics[height=0.13\columnwidth]{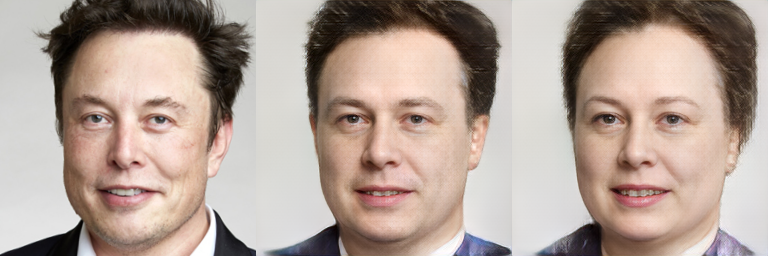}\\
               \includegraphics[height=0.13\columnwidth]{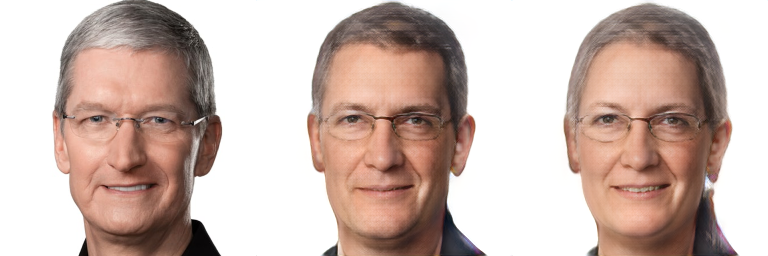}\\
               \includegraphics[height=0.13\columnwidth]{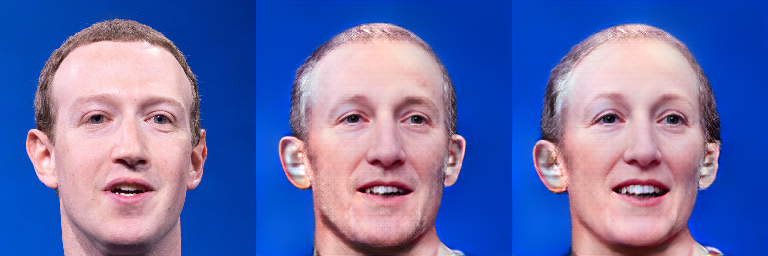}
               \caption{\small  Generated images pertaining to different gender for the same individual.}
               \label{fig:gender}
\end{figure*}

\begin{figure*}[!h]
        \centering
               \includegraphics[height=0.13\columnwidth]{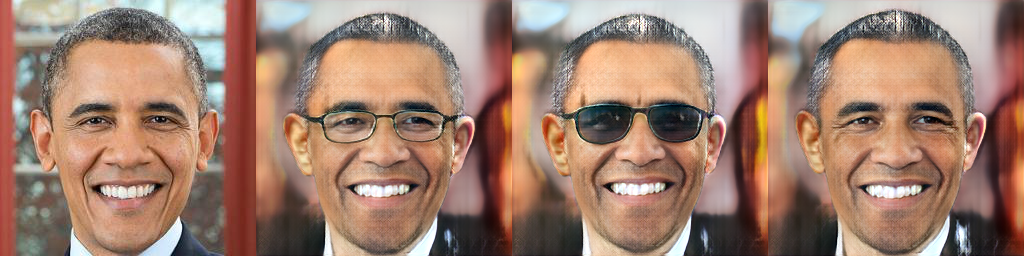}\\
               \includegraphics[height=0.13\columnwidth]{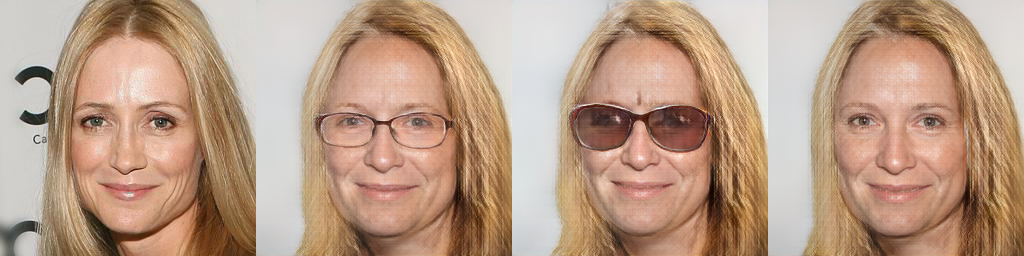}\\
               \includegraphics[height=0.13\columnwidth]{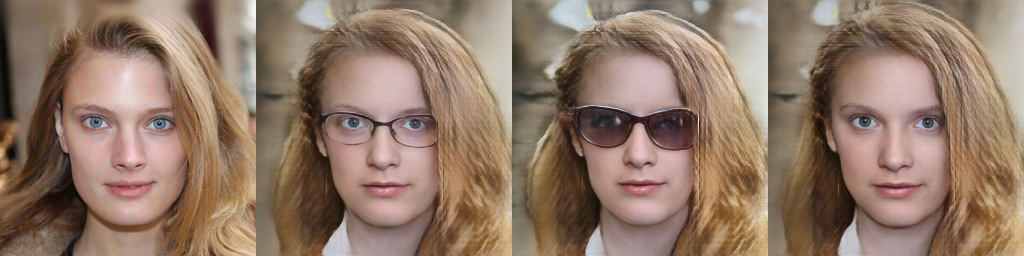}\\
               \includegraphics[height=0.13\columnwidth]{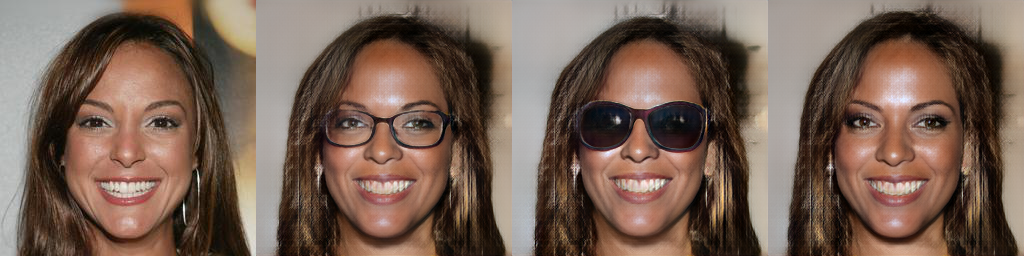}\\
               \includegraphics[height=0.13\columnwidth]{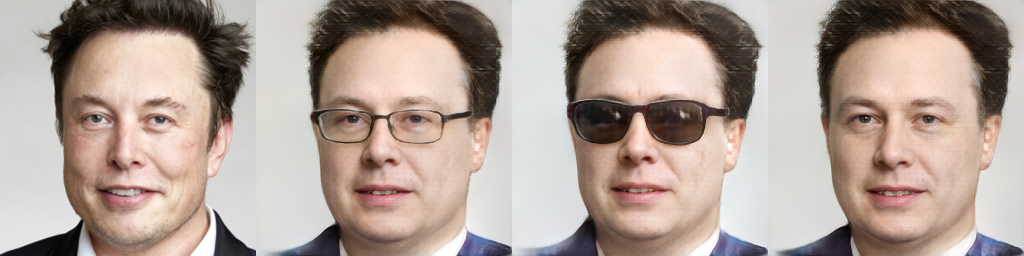}\\
               \includegraphics[height=0.13\columnwidth]{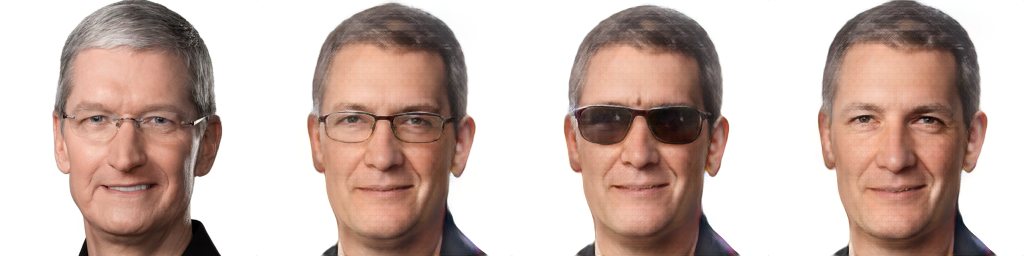}\\
               \includegraphics[height=0.13\columnwidth]{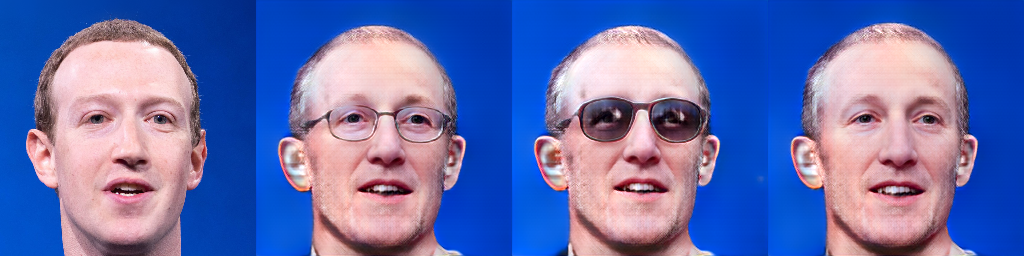}
               \caption{\small  Generated images pertaining to different level of ``glasses'' attribute for the same individual.}
               \label{fig:glasses}
\end{figure*}

\begin{figure*}[!h]
        \centering
               \includegraphics[height=0.13\columnwidth]{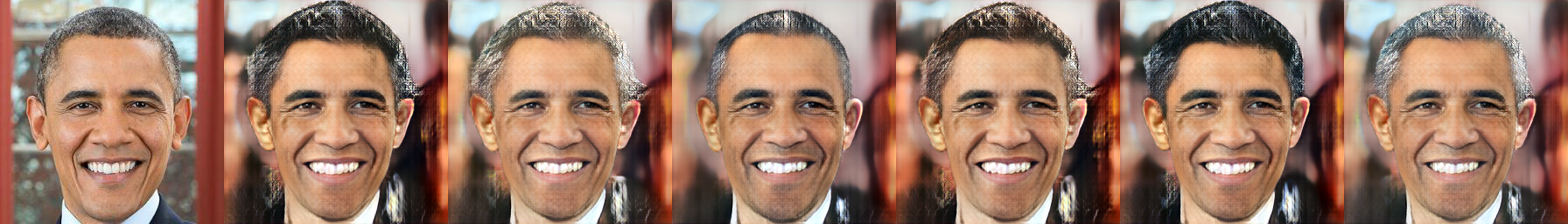}\\
               \includegraphics[height=0.13\columnwidth]{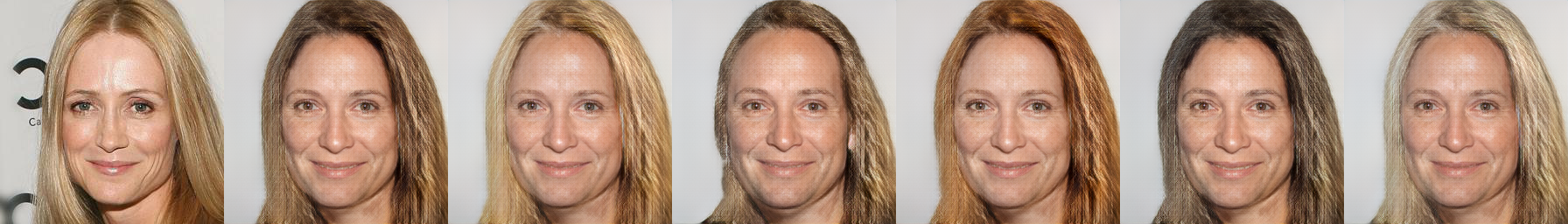}\\
               \includegraphics[height=0.13\columnwidth]{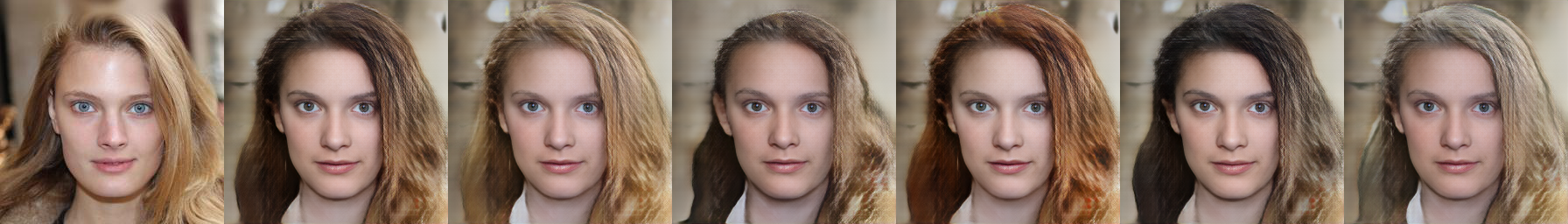}\\
               \includegraphics[height=0.13\columnwidth]{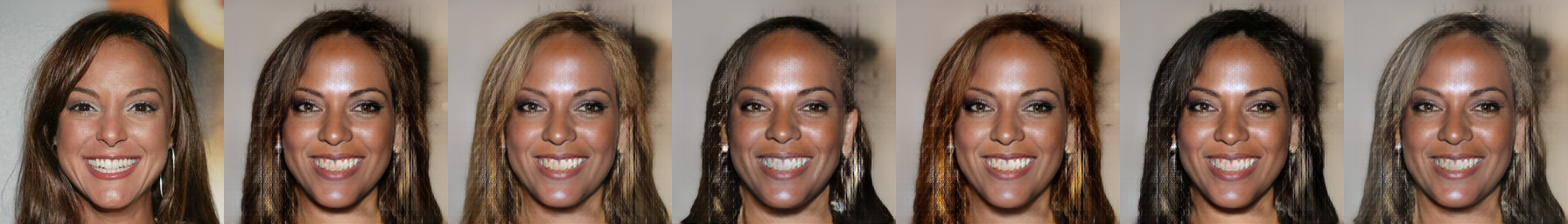}\\
               \includegraphics[height=0.13\columnwidth]{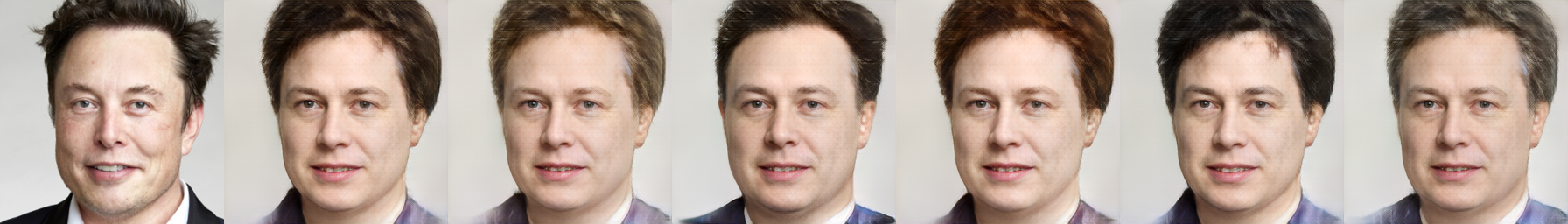}\\
               \includegraphics[height=0.13\columnwidth]{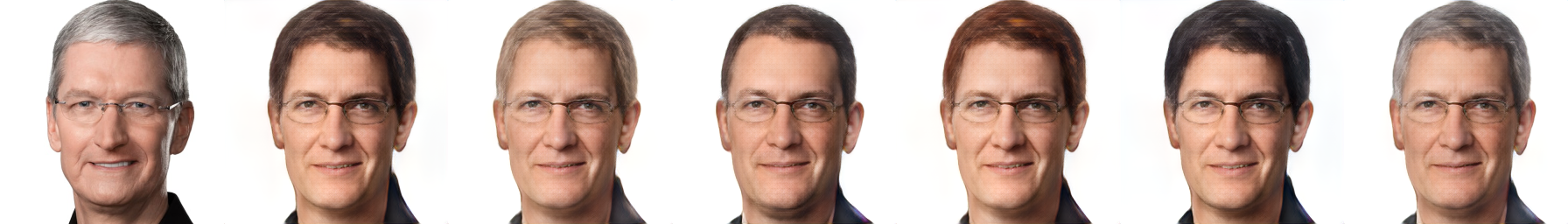}\\
               \includegraphics[height=0.13\columnwidth]{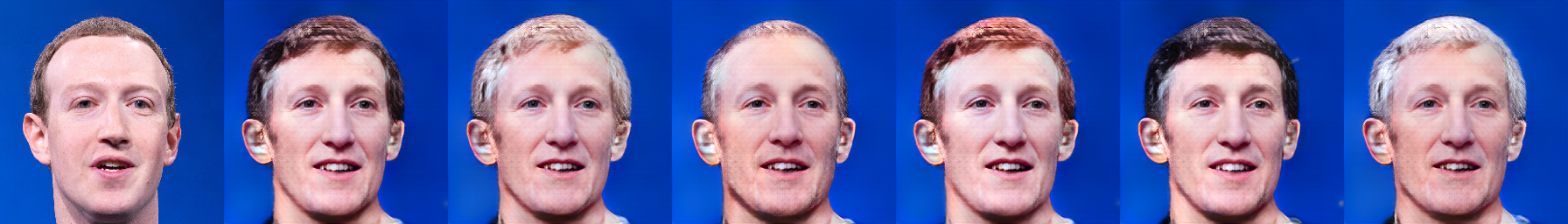}
               \caption{\small Generated images pertaining to different hair color for the same individual.}
               \label{fig:hair_color}
\end{figure*}

Some more generated images of the same individual are shown in Fig. \ref{fig:age}, \ref{fig:beard}, \ref{fig:ethnicity}, \ref{fig:gender}, \ref{fig:glasses}, and \ref{fig:hair_color}. We can see that our model can maintain most features in the image (keeps unchanged) and changes the attributes of interest separately. The results here are mostly qualitative.

\end{document}